\documentclass[journal]{IEEEtran}
\usepackage{hyperref}

\ifCLASSOPTIONcompsoc
\usepackage[nocompress]{cite}
\else
\usepackage{cite}
\fi

\ifCLASSINFOpdf
\else
\fi

\usepackage{algpseudocode,algorithm,algorithmicx}
\newcommand*\Let[2]{\State #1 $\gets$ #2}
\newcommand{\mL}{L\kern-0.12cm\char39}

\ifCLASSOPTIONcaptionsoff
\usepackage[nomarkers]{endfloat}
\let\MYoriglatexcaption\caption
\renewcommand{\caption}[2][\relax]{\MYoriglatexcaption[#2]{#2}}
\fi

\usepackage{url}

\usepackage{multirow}
\usepackage{mathtools}
\usepackage{amsthm}
\usepackage{amsfonts}
\newtheorem{definition}{Definition}

\DeclarePairedDelimiter\floor{\lfloor}{\rfloor}
\usepackage[dvipsnames]{xcolor}

\begin{document}
	
	\title{ Adaptive binarization based on fuzzy integrals }

	\author{Francesco~Bardozzo\IEEEauthorrefmark{1}\IEEEauthorrefmark{2}\IEEEauthorrefmark{5},
		Borja~De La Osa \IEEEauthorrefmark{3}\IEEEauthorrefmark{5},
		{\mL}ubom\'ira~Horansk\'a \IEEEauthorrefmark{4},
		Javier~Fumanal-Idocin \IEEEauthorrefmark{3},
		Mattia~delli~Priscoli\IEEEauthorrefmark{2},
		Luigi~Troiano\IEEEauthorrefmark{2},
		Roberto~Tagliaferri\IEEEauthorrefmark{2},~\IEEEmembership{Senior,~IEEE},
		Javier~Fernandez \IEEEauthorrefmark{3},
		Humberto~Bustince  \IEEEauthorrefmark{3},~\IEEEmembership{Senior,~IEEE}
		\IEEEcompsocitemizethanks{
			\IEEEcompsocthanksitem 	\IEEEauthorrefmark{1} Corresponding author F.Bardozzo, e-mail:  fbardozzo@unisa.it
			\IEEEcompsocthanksitem  \IEEEauthorrefmark{2} F. Bardozzo, M. Delli Priscoli, L. Troiano, R. Tagliaferri are with the DISA-MIS, University of Salerno, (Fisciano) SA, Italy.
			\IEEEcompsocthanksitem  \IEEEauthorrefmark{3} B. De La Osa, J. Fumanal-Idocin, J. Fernandez, H. Bustince are with Department of Statistics,	Computer Science and Mathematics, Public University of Navarra, Pamplona, Spain.
			\IEEEcompsocthanksitem \IEEEauthorrefmark{4} {\mL}.~Horansk\'a is with Institute of Information Engineering, Automation and Mathematics, Faculty of Chemical and Food Technology, Slovak University of Technology in Bratislava, Slovakia, e-mail: lubomira.horanska@stuba.sk.
			\IEEEcompsocthanksitem \IEEEauthorrefmark{5} Equal contribution.
		}
		\thanks{Manuscript received April XX, XXXX; revised August YY, YYYY.}}

	\markboth{Journal of \LaTeX\ Class Files,~Vol.~XX, No.~Y, Month~Year}%
	{Bardozzo \MakeLowercase{\textit{et al.}}:  Adaptive binarization based on fuzzy integrals}

	\maketitle

	\begin{abstract}
		Adaptive binarization methodologies threshold the intensity of the pixels with respect to adjacent pixels exploiting the integral images. In turn, the integral images are generally computed optimally using the summed-area-table algorithm (SAT). This document presents a new adaptive binarization technique based on fuzzy integral images through an efficient design of a modified SAT for fuzzy integrals. We define this new methodology as FLAT (Fuzzy Local Adaptive Thresholding).  The experimental results show that the proposed methodology have produced an image quality thresholding often better than traditional algorithms and saliency neural networks. We propose a new generalization of the Sugeno and $CF_{1,2}$ integrals to improve existing results with an efficient integral image computation. Therefore, these new generalized fuzzy integrals can be used as a tool for grayscale processing in real-time and deep-learning applications.	
	\end{abstract}


	%
	\IEEEpeerreviewmaketitle



	\section{Introduction}
	\label{sec:introduction}

	\IEEEPARstart{M}{ost} of the binary segmentation algorithms based both on deep learning (DL) or traditional models are built on taking advantage of the foreground/background recognition \cite{aich2018}.   Despite the multi-class semantic segmentation problems, the binary segmentation is specifically demanded in those applications where real-time performance is required and a simple but accurate structural and semantic representation is mandatory.
	In the literature, several image binarization algorithms based on both traditional and neural networks models are proposed for different applicative problems. For example, \emph{Cheremkhin et al.}\cite{cheremkhin2019comparative}  provide an extended review of traditional methodologies based on global and local binarization methods for hologram compression; \emph{Kalaiselvi et al.}\cite{kalaiselvi2017comparative} present a comparison between thresholding techniques for real-world and brain MRI image segmentation. Furthermore, \emph{Roy et al. }\cite{roy2014adaptive} provide a comparative study for the most common adaptive techniques.
	Recently, models based on convolutional networks are adopted for binarization and beyond.
	In particular, one of the natural evolutions of binarization approaches relies on the study of visual perception, better defined as visual saliency, and in the ability to distinguish and keep imprinted an object, a person or more generally a group of pixels on which the human attention is focused, both in the retina and in the post-processing phase, including the memorization step. \cite{achanta2009frequency}.
	At several level, the traditional approaches are embedded in DL models showing a fair balance between accuracy, generalization power and computational time costs \cite{dias2018using,dai2015convolutional, he2019deepotsu, fan2019road}. After all, the images are a matrix of values, thus enabling researchers to use binarization in complex networks \cite{yan2018weight}, Bayesian networks \cite{gross2019analytical} and biological networks/pathways \cite{bardozzo2018study}. Even if there are several binarization techniques in literature, none is the \emph{gold standard}.
	The traditional global thresholding algorithms are generally worse than the local ones. Moreover, combined models of local and global techniques process the same image several times showing a lack of performance over time \cite{sauvola2000adaptive, bradley2007adaptive}. \newline  Furthermore, the perturbations that could affect an image are heterogeneous (illumination changes, experimental noise, variable contrast, etc..) and depends on the represented subjects.  At several digital processing levels, such for example in the compressive sampling and lossy compression,  the different types and degrees of digital degradation could influence binarization accuracy. Also for this kind of problems, Information Theory provides quantization strategies, but at the cost of much greater estimation complexity \cite{goyal2008compressive}.  In genomic and proteomic analyses, the adaptive thresholding is exploited for the study of differential microarray spot intensities \cite{ wang2018automatic, hudaib2016new}. The objects analyzed in the images can be static or in motion, multiple or single. Traditional or neural-network-based binarization of real-world \cite{el2008fuzzy}, as well as, of micro-world \cite{boegel2015fully} could be used to establish relations between frames \cite{ciaparrone2019deep}. In this work, we focus our attention on local adaptive thresholding methods. In general, the latter are more accurate than the global ones and could be fine-tuned in an automatic way \cite{zemmour2019automatic, boegel2015fully}. \newline The idea behind the local adaptive thresholding relies on considering a threshold value for every pixel intensity or region of pixel intensities basing the analysis on its neighbouring pixels on a fixed or variable local window.  After all, the notion of adaptation has its roots in the concept of multi-scale analysis, structural variational analysis and representation of differential intensity values.
	As described by \emph{Bradley and Roth}\cite{bradley2007adaptive}, one of the most efficient local adaptive thresholding method comes from an extension of the Wellner's method \cite{wellner1993adaptive} and it is a generalized form of the \emph{Niblack} algorithm \cite{niblack1986introduction}. In particular, \emph{Bradley and Roth} adaptive thresholding method (known as \emph{Bradley} algorithm) exploits the representation power of integral images. Nevertheless, as proved in \emph{Debayle and Pinoli} \cite{debayle2009general}, the fuzzy integrals in the context of local adaptiveness show to outperform methodologies based on simple integral images. On the other hand, in the literature, there have already been attempts to modify the \emph{Bradley} algorithm. In particular, in these cases, a modification in the computation of the average neighbouring pixel intensities is used,  for example considering a weighted integral image \cite{wu2015handwritten}.  On the same line with the precedent authors, in this work, we propose a novel LAT, and we define it as \emph{FLAT}, which is the acronym of Fuzzy Local Adaptive Thresholding algorithm.  \emph{FLAT} is based on the logic of the \emph{Bradley} algorithm \cite{bradley2007adaptive}. In particular, \emph{FLAT} improves the thresholding accuracy leveraging a generalized form of the fuzzy integral images; for what is our knowledge, the latter approach has never been applied.
	The fuzzy integral images are computed from the integral images with a new efficient algorithm based on a modification of the summed-area-table algorithm (SAT) \cite{kasagi2014parallel} showing real-time performances (1100 fps over 200 $\times$ 200 pixels). The document is organized as follows:  the theoretical aspects of the three $FLAT $ variants  based on the generalizations of \emph{Sugeno} and $ CF_ {1,2} $ are explained in Section \ref{background}. Instead, in  Section \ref{adaptivefuzzyintegrals}, the new algorithms and changes to the $SAT$ algorithm are introduced.
	In Section \ref{res}, the results produced by our algorithms are compared to traditional and CNN-based adaptive approaches, both in terms of quality of the output and of performance. In particular, in the first sub-section \ref{comp1} the goodness of our algorithms is evaluated on a toy data set with controlled perturbations. Next, in sub-section \ref{comp3}, a larger data set of real world images portraying single/multiple objects ($\approx$ 2500 samples) is analyzed and the binarizations are compared. Finally, in  sub-section \ref{comp2}, our models, \emph{Bradley} algorithm and a state of the art CNN, in their optimal configurations, are compared on a dataset of $\approx$ 300 images hard to binarize. In conclusion, our 3 \emph{FLAT} algorithms show very accurate results and optimal performances.  Moreover, they appear to have better binarization capability than some state-of-the-art algorithms trained for convolutional networks.
	The implementation of 3 different variants of \emph{FLAT}, the pipeline and novel challenging datasets, are available at: \href{https://github.com/lodeguns/FuzzyAdaptiveBinarization}{https://github.com/lodeguns/FuzzyAdaptiveBinarization}.

	\section{Background}
	\label{background}
	\subsection{Fuzzy measures and fuzzy integrals}
	\label{gencho}
	Let $ n \in {\mathbb  N } $, $[n]=\{1,\dots,n\}$.
	A set function $\mu\colon 2^{[n]}\to [0,1]$ is a {\it fuzzy measure}, if the following conditions are satisfied:
	\begin{itemize}
		\item
		$\mu(a)\leq \mu(B)$ whenever $A\subseteq B$,
		\item
		$\mu(\emptyset)=0$, $\mu([n])=1$.
	\end{itemize}
	
	A fuzzy measure $\mu$ is {\it symmetric}, if for any $A,B\subseteq [n]$,  $|A|=|B|$ implies $\mu(A)=\mu(B)$ (here $|E|$ stands for the cardinality of the set $E$).   For example, the uniform fuzzy measure $\mu_{uni}$ given by
	\begin{equation}\label{uni}
	\mu_{uni}(E)=\frac{|E|}{n},
	\end{equation}
	for $E\subseteq [n]$, is symmetric.
	\newline
	\newline
	A function $A:[0,\infty[^n\to[0,\infty[$ is  an {\it aggregation function}, if $A$ is nondecreasing and
	$\inf\limits_{{\bf x}\in [0,\infty[^n} A({\bf x})=0,$
	$\sup\limits_{{\bf x}\in [0,\infty[^n} A({\bf x})=\infty.$
	
	An aggregation function $A:[0,\infty[^n\to[0,\infty[$ is
	\begin{itemize}
		\item  {\it internal}, if
		$\bigwedge\limits_{i=1}^n  x_{i}\leq A(x_1,\dots,x_n)
		\leq \bigvee\limits_{i=1}^n  x_{i}$, for each $(x_1,\dots,x_n)\in [0,\infty[^n$.
		
		\item {\it translation invariant}, if $A(x_1+c,\dots,x_n+c)=A(x_1,\dots,x_n)+c$, for all $c\in ]0,\infty[$
		and $(x_1,\dots,x_n)\in [0,\infty[^n$.
		\item {\it idempotent}, if $A(x,\dots,x)=x$, for each $x\in [0,\infty[$.
		\item {\it positively homogeneous}, if $A(c{\bf x})=cA({\bf x})$, for each ${\bf x}\in [0,\infty[^n$ and $c>0$.
		\item {\it comonotone additive}, if $A({\bf x}+{\bf \overline x})=A({\bf x})+A({\bf \overline x})$, for all comonotone vectors ${\bf x}, {\bf \overline x}\in [0,\infty[^n$ (vectors ${\bf x}=(x_1,\dots,x_n), {\bf \overline x}=(\overline x_1,\dots,\overline x_n)$ are
		comonotone, if $(x_i-x_j)(\overline x_i-\overline x_j)\geq 0$  for all $i,j \in \{1, \dots, n\}$).
		\item {\it comonotone maxitive (comonotone minitive)}, if $A({\bf x}\vee{\bf \overline x})=A({\bf x})\vee A({\bf \overline x})$
		($A({\bf x}\wedge{\bf \overline x})=A({\bf x})\wedge A({\bf \overline x})$), for all comonotone vectors ${\bf x}, {\bf \overline x}\in [0,\infty[^n$.
	\end{itemize}
	
	Let
	$\mu\colon 2^{[n]}\to [0,1]$ be a fuzzy measure.   The discrete {\it Choquet integral} with respect to the fuzzy measure $\mu$ is given by
	
	\begin{equation}\label{Choquet}
	Ch_\mu({\bf x})= \sum_{i=1}^n (x_{(i)}-x_{(i-1)})\cdot \mu(E_{(i)}),
	\end{equation}
	for any ${\bf x}=(x_1,\dots,x_n) \in [0,\infty[^n$, where $(\cdot)$ is a permutation on $[n]$ such that $x_{(1)} \leq \cdots \leq x_{(n)}$, with the convention $x_{(0)}=0$ and  $E_{(i)}=\{(i), \ldots , (n)\}$ for $i=1,\dots,n$.
	
	The {\it Sugeno integral} with respect to the fuzzy measure $\mu$ is given by
	\begin{equation}\label{Sugeno}
	Su_\mu({\bf x})= \bigvee\limits_{i=1}^n (x_{(i)} \wedge \mu(E_{(i)})),
	\end{equation}
	for  ${\bf x}=(x_1,\dots,x_n)  \in [0,\infty[^n$, with the same meaning of $x_{(i)}$ and $E_{(i)}$, $i=1,\dots,n$, as above.
	
	The  Choquet integral is an internal function, which is idempotent and  positively homogeneous and  gives back the considered fuzzy measure, i.e., $Ch_\mu({\bf 1}_E)=\mu(E)$
	for each $E\subseteq [n]$, where
	${\bf 1}_E$ stands for the indicator of the set  $E$.
	
	The Sugeno integrals is not bounded by the minimum from below, but it is bounded by the maximum from above. It is neither idempotent nor positively homogeneous
	(however,  the Sugeno integral is an idempotent, internal, positively homogenous function on the interval $[0,1]$).
	It  gives back the considered fuzzy measure, i.e., $Su_\mu({\bf 1}_E)=\mu(E)$, for each $E\subseteq [n]$.
	
	Moreover, the Choquet integral is  comonotone additive and translation invariant, while the Sugeno integral is comonotone maxitive and comonotone minitive
	(for more details see, e.g., \cite{grabisch2009aggregation}).
	
	\subsection{Generalized Sugeno integral}
	\label{gensug}
	We modify formula (\ref{Sugeno}) defining Sugeno integral by replacing maximum and minimum operators by some more general functions. The obtained functional  can be regarded as a generalization of the Sugeno integral.
	
	\begin{definition}{
			\label{Adef}
			Let $\mu\colon 2^{[n]}\to [0,1]$ be a symmetric fuzzy measure,
			$F\colon [0,\infty[\times [0,1]\to [0,\infty[$ be  a binary  function,
			$G\colon [0,\infty[^n\to [0,\infty[$ be  an $n$-ary  function.
			A {\it Sugeno-like $FG$-functional} is a function $A \colon [0,\infty[^n\to [0,\infty[$ given by
			\begin{equation}{\label{A}
				A(x_1,\dots,x_n)=G\left(F(x_{(1)},\mu(E_{(1)})),\dots,F(x_{(n)},\mu(E_{(n)}))\right), 
			}
			\end{equation}
			for  ${\bf x}=(x_1,\dots,x_n)  \in [0,\infty[^n$, with the same meaning of $x_{(i)}$ and $E_{(i)}$, $i=1,\dots,n$, as above.}
	\end{definition}
	
	The correctness of the definition depends on  whether the functional $A$  given by formula (\ref{A}) gives back the same value if some ties occur in a  vector ${\bf x} $ and there is more than  one permutation ordering this vector nondecreasingly.
	The symmetry of the fuzzy measure $\mu$ considered in Definition \ref{Adef} ensures that functional $A$ is well-defined.
	In fact, for particular cases of $G$, assumptions under which $A$ is well-defined can be weakened. For example,
	the case of $G$ being the maximum operator and $F$ an arbitrary fusion function was deeply studied in \cite{horanska2018generalization},
	wherein assumptions under which $A$ is well-defined for an arbitrary  fuzzy measure $\mu$ and a complete characterization of the functional $A$ and its properties can be found.
	
	The following three instances of Sugeno-like $FG$-functionals are of particular interest for us:
	\begin{itemize}
		\item[(i)]
		Let $G(x_1,\dots,x_n)=\bigvee\limits_{i=1}^n x_i$ and $F(x,y)=x\wedge y$.
		Then we get
		\begin{equation}\label{A_1}
		A_1({\bf x})=\bigvee\limits_{i=1}^n \left( x_{(i)}\wedge \mu(E_{(i)})\right),
		\end{equation}
		so we recover the Sugeno integral, i.e. $A_1=Su$.
		\item[(ii)]
		Let $G(x_1,\dots,x_n)=\sum\limits_{i=1}^n x_i$ and $F(x,y)=x\cdot y$.
		Then we obtain
		\begin{equation}\label{A_2}
		A_2({\bf x})=\sum\limits_{i=1}^n \left( x_{(i)}\cdot \mu(E_{(i)})\right).
		\end{equation}
		\item[(iii)]
		Let $G(x_1,\dots,x_n)=\sum\limits_{i=1}^n x_i$ and $F(x,y)=\frac{xy}{x+y-xy}$.
		Then we obtain
		\begin{equation}\label{A_3}
		A_3({\bf x})=\sum\limits_{i=1}^n \frac{ x_{(i)}\cdot \mu(E_{(i)})}{x_{(i)}+ \mu(E_{(i)})-x_{(i)}\cdot \mu(E_{(i)})}.
		\end{equation}
		Note, that $F$ is the Hamacher t-norm corresponding to the parameter $\lambda=0$.
	\end{itemize}
	
	A straithforward computation gives us the following properties of $A_2$ and $A_3$: Both $A_2$ and $A_3$ are aggregation functions, since they are nondecreasing and
	\begin{equation}
	\begin{multlined}
	\ \ \ \ \inf\limits_{{\bf x}\in [0,\infty[^n} A_2({\bf x})=\inf\limits_{{\bf x}\in [0,\infty[^n} A_3({\bf x})=0, \\
	\sup\limits_{{\bf x}\in [0,\infty[^n} A_2({\bf x})=\sup\limits_{{\bf x}\in [0,\infty[^n} A_3({\bf x})=\infty. \  \ \
	\end{multlined}
	\end{equation}

	Both $A_2$ and $A_3$ are bounded by the minimum from below, but  not  bounded by the maximum from above.\newline
	$A_2$ is positively homogeneous,  but $A_3$ is not.
	Neither $A_2$ nor $A_3$ are idempotent, giving back the capacity, comonotone additive, comonotone maxitive, translation invariant.
	\newline Finally, we define $A_4 = Ch$ in order to keep uniformity of  the notation in the following paragraphs .

	\subsection{Computation of the integral image $S$ with SAT}
	\label{brad}
	
	Let $ n,m \in {\mathbb  N } $, $[n]=\{1,\dots,n\},\, [m]=\{1,\dots,m\}$.
	An original image $I$ consisting of $n\times m$ pixels arranged in $n$ rows and $m$ columns  is associated with the  matrix
	$\left (p(x,y)\right)_{(x,y)\in [n]\times[m]}$ assigning the intensity $p(x,y)$ to the each pixel $(x,y)\in [n]\times[m]$.
	
	In the \emph{Bradley} algorithm, the binarized pixel values are determined considering the average pixel intensities $p_a$ of its neighbouring pixels. The  central role in determining the value of $p_a$ is played by the so-called integral image.  The \emph{integral image} $S$ is the matrix $\left(S(x,y)\right)_{(x,y)\in [n]\times[m]}$, defined for any pixel  $(x,y)\in [n]\times[m]$  by the following formula \ref{eq1}:
	\begin{equation}
	\label{eq1}
	S(x,y) =\sum_{i \leq x }\sum_{j \leq y } p(i,j).
	\end{equation}
	Determination of $S$ has time complexity of $O(l* (n*m))$, which is derived by the $l$ number of times that the Equation above is applied. However, leveraging the summed-area table algorithm (SAT) \cite{crow1984summed}, the  computation of $S(x,y)$ can be maintained constant and  the SAT time complexity remains fixed to $O(n*m)$. The SAT can be developed efficiently computing for each pixel $(x,y)\in [n]\times[m]$ the column-wise prefix-sums and the row-wise prefix-sums \cite{kasagi2014parallel}, as it is shown in Equation \ref{eq2}:
	\begin{equation}
	\label{eq2}
	\begin{multlined}
	S(x,y)=p(x,y) + S(x, y-1) + \\ S(x-1, y) - S(x-1, y-1),
	\end{multlined}
	\end{equation}
	with convention $S(0,k)=0$, for each $k=0,\dots,m$ and $S(l,0)=0$, for each $l=0,\dots,n$.
	
	\subsection{\emph{Bradley} algorithm based on the integral image $S$}
	\label{brad2}
	Let us denote by $[(x_1, y_1),(x_2, y_2)]$ the rectangle determined by the upper left corner $(x_1, y_1)$ and  the lower right corner $(x_2, y_2)$.
	Once $S$ is obtained, the sum of the pixel intensities in a rectangle $[(x_1, y_1),(x_2, y_2)]$ denoted by $p_s(x_1,y_1,x_2,y_2)$,  is given by Equation \ref{eq3}:
	\begin{equation}
	\label{eq3}
	\begin{multlined}
	p_s(x_1,y_1,x_2,y_2) = S(x_2, y_2) - S(x_2, y_1) - \\ S(x_1, y_2) + S(x_1, y_1),
	\end{multlined}
	\end{equation}
	where $1\leq x_1\leq x_2\leq m$, $1\leq y_1\leq y_2\leq n$.  Thus, the average value of the pixel intensities in the rectangle $[(x_1, y_1),(x_2, y_2)]$,  is given by Equation \ref{eq4}:
	\begin{equation}
	\label{eq4}
	p_a(x_1,y_1,x_2,y_2) = \frac{p_s(x_1,y_1,x_2,y_2)}{(x_2-x_1)\times(y_2 - y_1)}
	\end{equation}
	with $1\leq x_1\leq x_2\leq m$, $1\leq y_1\leq y_2\leq n$.
	In the second part of the process, the pixel intensity  of the original image $I$ is compared pixel-by-pixel with the average value of the pixel intensities in the local window around the current pixel.
	For a given size $2s\times 2s$ of the local window, the \emph{Bradley} algorithm iteratively binarizes the original image and provides the binary values $I_b(x,y)$ for each pixel $(x,y)$,  as described in the following formula:
	\begin{equation}
	\label{eq5}
	I_b(x,y) = \left\{
	\begin{array}{ll}
	1 &\mbox{ if } \ p(x,y) \leq p_a(x_1,y_1,x_2,y_2)\times t, \\[.2em]
	0 &\mbox{otherwise},
	\end{array}
	\right.
	\end{equation}
	where $(x_1,y_1,x_2,y_2)=(x-s,y-s,x+s,y+s)$. Note that the local window needs to be changed, if it is not within the borders of the original image.
	Note also that it is considered only a percentage of $p_a$ controlled by the sensitivity value \emph{t}, which is defined in the interval $[0,1]$.
	\begin{figure}[!hpb]
		\centering
		\includegraphics[width=84mm]{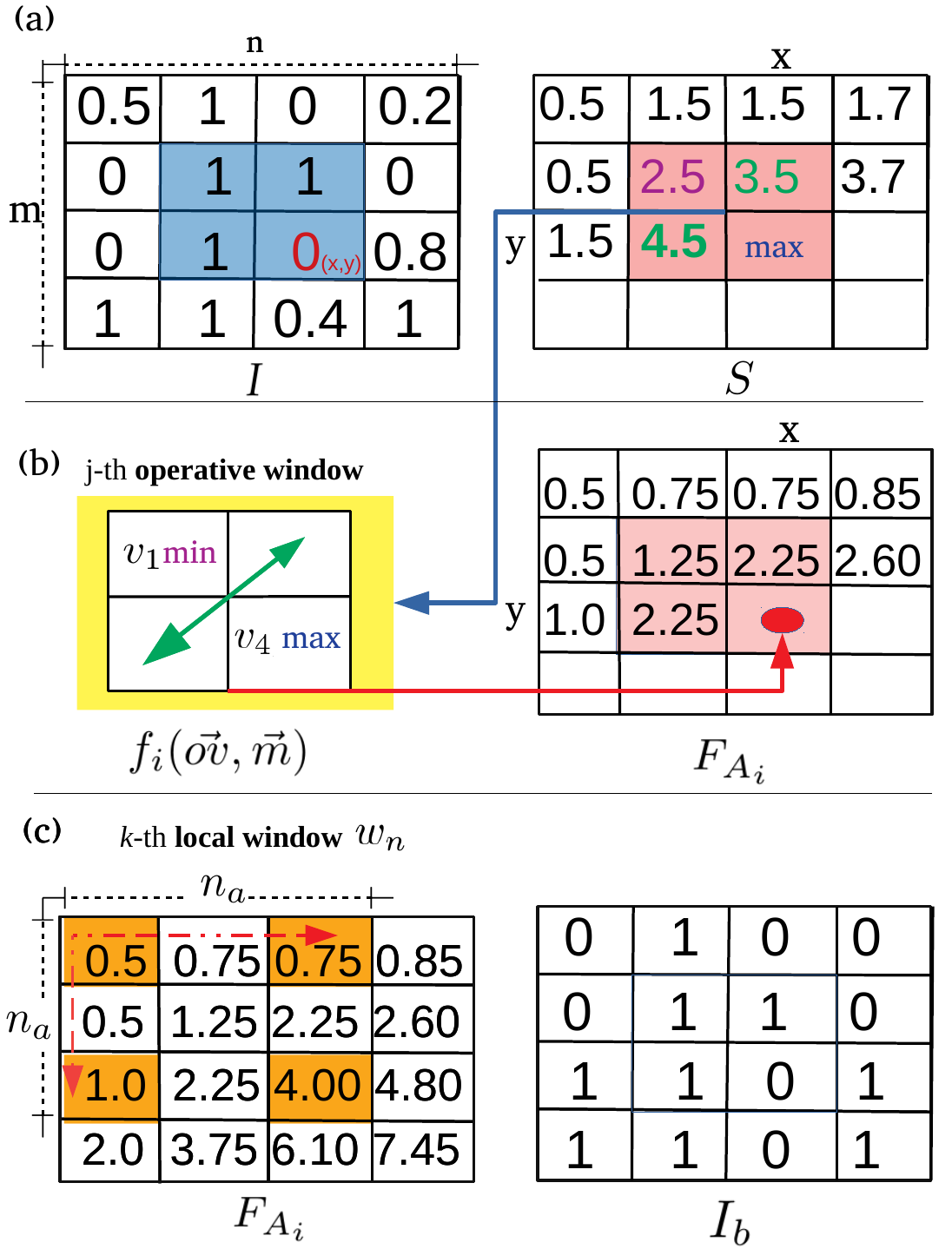}
		\caption{ Here a whole overview of the \emph{FLAT} algorithm.    \label{fig_000} In Figure \ref{fig_000} - Boxes \textbf{(a-b)}, the steps of Algorithm \ref{alg:algo1} for the computation of both the integral image $S$ and the fuzzy integral image $F_{A_i}$ are shown (see section \ref{brad} and \ref{flat1}).  The blue square in Box \textbf{(a)} represents the current value of $p(x,y) \in I$ (in red) and its neighboring pixels defined in the $j$-th operative window. For each pixel $p(x,y)$ and for every  \emph{j}-th operative window (yellow box), the 4 values in $S$ are mapped with the associated fuzzy measures through $F_{A_i}(x,y)=f_i(\vec{ov}, \vec{m})$ for $i=1,2,3,4$. The output of the fuzzy-based integral functional computation is saved in the fuzzy integral image $F_{A_i}$. This computation is described in formula \ref{fuz}. In the \emph{j}-th operative window, the fixed values of min $v_1$ and max $v_4$ are represented in violet and blue. For the decision of $v_2$ and $v_3$, the green arrow indicates the $P_{swap}$ action as described in Procedure \ref{eq6}. In Figure \ref{fig_000} Box \textbf{(c)}, the \emph{k}-th local search window $w_n$ used for the locally adaptive thresholding is shown. It is important to underline that, as it is described in Algorithm \ref{alg:algo2}, only the 4 values in the orange rectangles are used for the binarization. These 4 values are not necessarily adjacent like in the operative window. The dashed red arrows show the local window sliding directions, from up to down, from left to right. The local window has a fixed size of $n_a * n_a$. The $I_b$ indicates the binarized image given in output considering the $b$-type fuzzy integral image. }
	\end{figure}

	\section{The \emph{FLAT} algorithm}
	\label{adaptivefuzzyintegrals}
	The \emph{FLAT} algorithm is described in the following 2 sub sections. In the former, it is shown how the generalized fuzzy integrals are combined with the calculation of the integral image, demonstrating also why the computational complexity remains the same as the traditional SAT algorithm (see Algorithm \ref{alg:algo1}). In the latter, the binarization is applied according to the fuzzy integral images (see Algorithm \ref{alg:algo2}).
	
	\subsection{Fuzzy integral image computation ($\mathbf{F_{A_i}}$):}
	\label{flat1}
	Fuzzy integrals are used to avoid uncertainty in binarization and beyond, showing various application fields in the most different research areas \cite{el2008foreground, barreto2018fuzzy}. The main disadvantage of fuzzy integrals, like the Choquet integrals,  relies in allocating further computational effort to the element sorting, in order to respect the monotonicity property (see section \ref{background}). Despite this last observation and looking closely at the cascade construction of an integral image, in a constant sorting time, it is possible to adopt the procedure applied in the SAT and optimally generate the fuzzy integral image $F$.
	As shown in Figure \ref{fig_000} - Box \textbf{(a-b)}, once the integral image S is computed (see subsection \ref{brad}), for each pixel $(x,y)$ in the $j$-th operative window, it is possible to compute the \emph{fuzzy integral image} $F_{A_i}$ , as follows:
	\begin{equation}
	\label{fuz}
	\begin{multlined}
	F_{A_i}(x,y)=\\
	A_i(S(x,y),S(x, y-1),S(x-1, y),S(x-1, y-1)),
	\end{multlined}
	\end{equation}
	where $A_i\colon[0,\infty[^4\to[0,\infty[$, for $i=1,2,3,4$,  is  one of the fuzzy integral-based functionals mentioned in the previous section, namely  the Sugeno integral $A_1=Su$,
	the Sugeno-like $FG$-functionals $A_2$ and $A_3$, respectively and the Choquet integral $A_4=Ch$. As a fuzzy measure we adopt the uniform fuzzy measure $\mu_{uni}$ defined by formula (\ref{uni}).

	As shown in Algorithm \ref{alg:algo1}, the procedure takes advantage of the natural ordering of the four elements aggregated in  (\ref{fuz}), obtaining a vector of ordered values $\vec{ov}$ and an associated static vector of fuzzy measures $\vec{m}$.
	
	In fact,  the maximum value $v_4$ is the element $S(x,y)$ present in the right lower corner of the $j$-th operative window, while, the minimum value $v_1$ is  $S(x-1, y-1)$ element ((see Figure \ref{fig_000} - Box \textbf{(b)}) - $min$ in violet, $max$ in blue). In order to complete the sorting, we need just  eventually to swap $s_1=S(x, y-1)$ and $s_2=S(x-1, y)$ by the following swap operation $P_{swap}$ ((see Figure \ref{fig_000} - Box \textbf{(b)}) - double green arrow):
	
	\begin{equation}
	\label{eq6}
	P_{swap}(s_1,s_2) =
	\left\{\begin{matrix}
	v_2 = s_1 , v_3 = s_2  \ \ \ \ \ \ if \ s_1 < s_2
	\\
	v_2 = s_2 , v_3 = s_1 , \ \ \ \ otherwise
	\end{matrix}\right.
	\end{equation}
	Thus, the final result is a one dimensional array of sorted values: $\vec{ov} = [v_0, v_1, v_2, v_3, v_4]$, where by convention $v_0 = 0$.
	Moreover, since $E_{(i)} = \{(1), \dots (4)\}$ is the subset of indices of the $4-i+1$ greatest component of $\vec{ov}$, for the uniform fuzzy measure $\mu_{uni}$ defined by formula (\ref{uni}), we have $\mu_{uni}(E_{(i)})=\frac{4-i+1}{4}$.
	Hence, we deal always with the same vector of fuzzy measures $\vec{m} = [\mu_{uni}(E_{(1)}),\mu_{uni}(E_{(2)}),\mu_{uni}(E_{(3)}),\mu_{uni}(E_{(4)})]=[1, \ \ 0.75, \ \ 0.50,\ \  0.25]$.
	In Algorithm \ref{alg:algo1}, a bridge function $f_i(\vec{ov}, \vec{m})$ for each pixel $p(x,y)$is defined, in order to map each fuzzy integral-based functional computation (for $i=1,2,3,4$) on the two vectors: $\vec{ov}$ and $\vec{m}$ for the $j$-th operative window. As shown in Figure \ref{fig_000}, the fuzzy integral image $F_{A_i}$ could be computed with different bridge functions  $f_i(\vec{ov}, \vec{m})$, varying only the values of $\vec{ov}$ and $\vec{m}$ and maintaining the algorithmic structure unchanged. Only the 4 operative window corners are considered at a time ($O_{f_{i}}(4)$) leaving the computational complexity polynomial in time, as it is for the orginal SAT ($O(n \times m) + O_{f_{i}}(4) + \dots  =  O(n \times m)$).

	\subsection{Adaptive binarization with the $\mathbf{F_{A_i}}$:}
	Algorithm \ref{alg:algo1} outputs the fuzzy integral image $F_{A_i}$. Then, the latter is given in input to Algorithm \ref{alg:algo2} for binarization. For what is concerning the binarization, $F_{A_i}$ will be leveraged as $S$ is exploited in \emph{Bradley} algorithm (see section \ref{brad}). However, in Algorithm \ref{alg:algo2} a modified version of the \emph{Bradley} algorithm is presented according to our constraints. In detail, $F_{A_i}$ is computed with the different integral generalization presented in sections \ref{gencho} and \ref{gensug}. Furthermore, the size and the coordinates of the sliding nearest neighbor's pixel local window $w_n$ is set with the dimensional parameter $n_a$. The latter is computed through 2 empirical parameters: $ a_1 $ and $ a_2 $. As it is described above, the $w_n$ is used to locally binarize the central pixels. Thus, $w_n$ is sized and positioned following the procedure described in Algorithm \ref{alg:algo2}.  The area of $w_n$ is equal to $n_a^2$. The dimensional parameter $n_a$ is computed as follows:
	
	\begin{equation}
	\label{eq8}
	n_a = \floor*{\frac{ min(n,m) }{a_1 \times a_2} }.
	\end{equation}
	and it is based on the $(n,m)$ dimensions of $I$. The parameters $ a_1 $ and $ a_2 $, as well as $t$, can be varied iteratively to improve the accuracy until the binarized image at the $optimum$ ( $I_b^*$) is found. In particular, the $I_b^*$ indicates the $optimum$ binarization in terms of the best $F_m$ value \cite{ntirogiannis2008objective} with respect to the ground truth. The subscript $b$ indicates which method of binarization is applied, such as, for example, if we consider $b$ equal to $F_{A_1}$, $I_{F_{A_1}}$ is the original image binarized with the Sugeno integral image and $I^*_{F_{A_1}}$ is its binarization at the $optimum$.
	
	\begin{algorithm}[t]
		\caption{\label{alg:algo1}: Computation of $F_{A_i}$. (\emph{FLAT} - Step 1)}
		\begin{algorithmic}
			\Require{Gray-scale image \emph{I} with intensities in $[0,1]$. }
			\Function{Flat- $F_{A_i}$ }{$I$}
			\Let{$n, m$}{$dim(I)$} \Comment{Dimension of \emph{I} }
			\Let{$S$}{allocate a zero-matrix with size $(n,m)$}
			\Let{$F_{A_i}$}{allocate a zero-matrix with size $(n,m)$}
			\Let{$\vec{m}_1$}{$[1, 0.75, 0.50, 0.25]$}
			\Let{$\vec{m}_2$}{$[1,  0.50 ]$}
			\For{$r \gets 1 \textrm{ to } n$}
			\For{$c \gets 1 \textrm{ to } m$}
			\If{$r \neq 1 \wedge c \neq 1$}
			\Let{$v_1$}{$S[r - 1,c - 1]$}
			\Let{$s_1$}{$S[r,c - 1]$}
			\Let{$s_2$}{$S[r - 1,c] $}
			\Let{$S[r,c]$}{$I[r,c] + s_1 + s_2 - v_1$}
			\Let{$v_4$}{$S[r,c]$}
			\Let{$v2, v3$}{$P_{swap}(s_1, s_2)$ - Procedure \ref{eq6}}
			\Let{$v_0$}{$0$}
			\Let{$\vec{ov}$}{$[v_0,v_1,v_2,v_3,v_4]$}
			\Let{$F_{A_i}[r,c]$}{$f_{i}(\vec{ov}, m_1)$}  \Comment{$i=1,2,3,4$}
			\ElsIf{$r \neq 1$}
			\Let{$v_1$}{$S[r - 1,c] $}
			\Let{$S[r,c]$}{$I[r,c] + v_1$}
			\Let{$v_4$}{$S[r,c]$}
			\Let{$\vec{ov}$}{$[v_0,v_1,v_4]$}
			\Let{$F_{A_i}[r,c]$}{$f_{i}(\vec{ov}, m_2)$} \Comment{$i=1,2,3,4$}
			\ElsIf{$c \neq 1$}
			\Let{$v_1$}{$S[r,c-1] $}
			\Let{$S[r,c]$}{$I[r,c] + v_1$}
			\Let{$v_4$}{$S[r,c]$}
			\Let{$\vec{ov}$}{$[v_0,v_1,v_4]$}
			\Let{$F_{A_i}[r,c]$}{$f_{i}(\vec{ov}, m_2)$}  \Comment{$i=1,2,3,4$}
			\Else
			\Let{$S[r,c]$}{$I[r,c]$}
			\Let{$F_{A_i}[r,c]$}{$S[r,c]$}
			\EndIf
			\EndFor
			\EndFor
			\State \Return{$F_{A_i}$}
			\EndFunction

		\end{algorithmic}
	\end{algorithm}
	\begin{algorithm}[t]
		\caption{\label{alg:algo2}: Binarization based on  $F_{A_i}$ (\emph{FLAT} - Step 2) }
		\begin{algorithmic}
			\Require{Gray-scale image \emph{I} with intensities in $[0,1]$.}
			\Require{ The fuzzy integral image $F_{A_i}$.}
			\Require{ The parameters $a_1$ and $a_2$.  }
			\Require{ The sensitivity parameter $t$ }
			\Function{Flat- $I_b$ }{$I, F_{A_i}, a_1, a_2, t$}
			\Let{$n, m$}{$dim(I)$} \Comment{Dimension of \emph{I} }
			\Let{$I_b$}{allocate a zero-matrix with size $(n,m)$}
			
			\Let{$n_a$}{Defined in Formula \ref{eq8} with $a_1$ and $a_2$ }
			
			\For{$r \gets 1 \textrm{ to } n $}
			\For{$c \gets 1 \textrm{ to } m$}
			\Let{$y_0$}{$max(r-n_a, 0))$} \Comment{Set the $w_n$}
			\Let{$y_1$}{$min(r+n_a, r)$}
			\Let{$x_0$}{$max(c-n_a, 0)$}
			\Let{$x_1$}{$max(c+n_a, c)$}
			\Let{$p_{area}$}{$(y_1-y_0)*(x_1-x_0)$}
			\Let{$p_s$}{$F_{A_i}[y_1,x_1]$-$F_{A_i}[y_0, x_1]$-$F_{A_i}[y_1, x_0]$+ $F_{A_i}[y_0, x_0]$}
			
			\Let{$p_a$}{$\frac{p_s}{p_{area}}\newline$}
			\If{$I[r,c] \leq p_a \times (1-t)$}
			\Let{$I_b[r,c]$}{1}
			\Else
			\Let{$I_b[r,c]$}{0}
			\EndIf
			\EndFor
			\EndFor
			\State \Return{$I_b$}
			\EndFunction

		\end{algorithmic}
	\end{algorithm}

	\begin{figure}[t]
		\centering
		\begin{tabular}{@{\hspace{-0.5\tabcolsep}}c @{\hspace{-0.2\tabcolsep}}c@{\hspace{1\tabcolsep}} | c @{\hspace{-0.2\tabcolsep}}c@{\hspace{-0.5\tabcolsep}}  } \\
			Image \textbf{a} &
			Ground truth  &
			Image \textbf{c} &
			Ground truth
			\\
			\frame{\includegraphics[width=20mm]{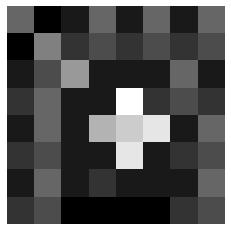}}&   \frame{\includegraphics[width=20mm]{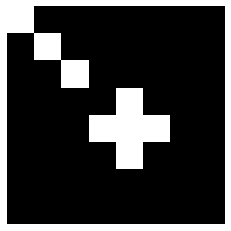}} &
			\frame{\includegraphics[width=22mm]{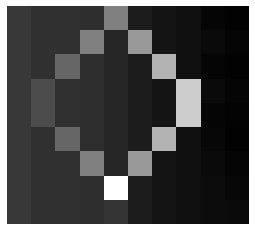}}&   \frame{\includegraphics[width=22mm]{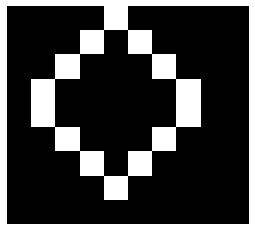}}
			\\[-2pt]
			$I_B$ & $\mathbf{I_{A_2}^*}$-$CF_{1,2}$ &   $I_B$  & $\mathbf{I_{A_2}^*}$-$CF_{1,2}$
			\\[-0.2pt]
			\frame{\includegraphics[width=20mm]{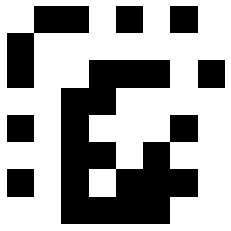}} &   \frame{\includegraphics[width=20mm]{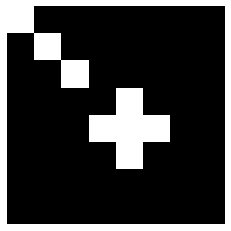}}  &
			\frame{\includegraphics[width=22mm]{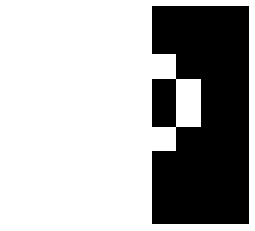}}&   \frame{\includegraphics[width=22mm]{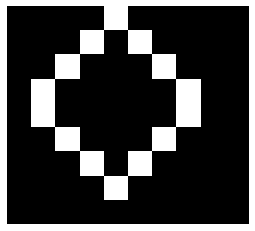}}
			\\[-2pt]
			$I_B$  &   $I_{A_3}$-Hamacher    &   $I_B$  & $\mathbf{I^*_{A_3}}$-Hamacher
			\\[-0.2pt]
			\frame{\includegraphics[width=20mm]{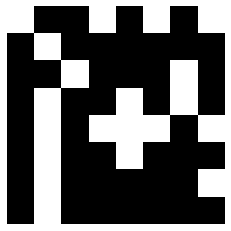}} &   \frame{\includegraphics[width=20mm]{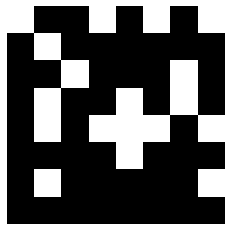}} &
			\frame{\includegraphics[width=22mm]{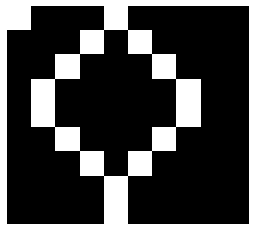}}&   \frame{\includegraphics[width=22mm]{imgs/1ff}}
			\\[-2pt]
			$I_B$  &	 $I_{A_4}$-Choquet   &  $I_B$  & $I_{A_4}$-Choquet
			\\[-0.2pt]
			\frame{\includegraphics[width=20mm]{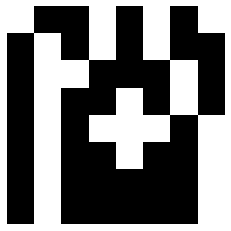}} &   \frame{\includegraphics[width=20mm]{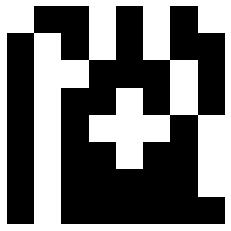}} &
			\frame{\includegraphics[width=22mm]{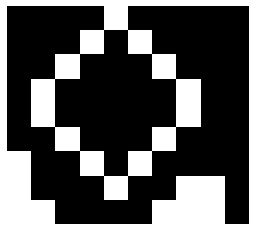}}&   \frame{\includegraphics[width=22mm]{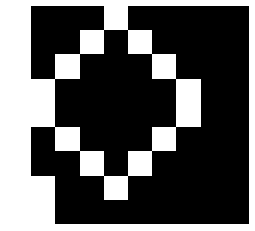}} 		
			\\    
		\end{tabular}
		\caption{ 	\label{fig_01}
			In  Figure \ref{fig_01}, the binarization results $I_b$ of two images are shown. In particular, Image \textbf{a} and Image \textbf{c}  of the $toy$ dataset. The binarization results based on our methods (respectively $I_{b}$, for $b=A_2, A_3, A_4 $) and the \emph{Bradley} algorithm ($I_B$) are compared, with the same parameter configurations of $w_n$ and $t$ (for numerical details see Table \href{https://github.com/lodeguns/FuzzyAdaptiveBinarization}{S-1}). The best values for the $SSIM$ are indicated with an asterisk and in bold. The images, as it is described in  section \ref{comp1} and it is showed in detail in Table \ref{per_table}, present different types of perturbations, that differently affect an accurate binarization.
		}
		
	\end{figure}

	\subsection{Dataset}
	\label{datasets}
	
	In order to test and compare our algorithms, we leverage a controlled toy dataset and a saliency \emph{MSRA-B} dataset  \cite{jiang2013salient}. These datasets are provided with ground truths (GTs) and have the following characteristics: \newline
	\textbf{Toy dataset:} The toy dataset is a novel challenging set of 8 images in which are applied several types of perturbations.  In particular, images are labeled alphabetically from \textbf{a} to \textbf{h}. These challenging images have a very small size with odd and even dimension  ($9 \times 9$ and $8 \times 8$ pixels). The pixel intensity is normalized in the interval $[0,1]$ and with a decimal precision of $0.01$.  In particular, these images have been designed in a methodological way to present increasing levels of difficulty for the binarization. Furthermore, the odd and even sizes are suitable for testing the correct sliding of the local window $w_n$. In particular, the dataset is built with an accurate modification of the pixel intensities with respect to the interplay of 5 specific challenging characteristics. Thus, the design of the images reflects 5 challenges: high-low contrast variations ($ \gamma_0 $), spatial variations in lighting ($ \gamma_1 $), additive random noise ($ \gamma_2 $), motifs of  structured noise ($ \gamma_3 $), smoothed borders ($ \gamma_4 $). Moreover, In Table \ref{per_table} the percentage of extension of the applied perturbations and the variability in intensity between the maximum and minimum average perturbation intensity are shown.
	\newline
	\textbf{Test set:}
	The second dataset comes from an accurate selection of $5.000$ images collected from the \emph{MSRA-B} dataset \cite {jiang2013salient}. This dataset is used for saliency analyses and the GTs are suited for testing saliency foreground/background extraction. Thus, $2413$ images are selected from \emph{MSRA-B} applying a global threshold filtering (\emph{Otsu} method \cite{otsu1979threshold}). In particular, the \emph{Otsu} predicted masks are compared with the GTs and only the original images with an $F_1$ measure greater than or equal to $0.7$ are selected.
	This guarantees to make fair comparisons between the binarizations/predictions made by \emph{DSS-Net} \cite{hou2017deeply} (see also Section ref) and those obtained by traditional algorithms and our fuzzy algorithms.

	\begin{table}[!htbp]
		\caption{Summary of perturbations for the toy dataset.} \label{per_table}
		\begin{tabular}{cccc}
			Image                & Perturbation            & Coverage       & Intensity Variability \\
			\textbf{a}           & $\gamma_0$              & $\approx 87\%$ & 0.20                   \\
			\textbf{}            & $\gamma_3$              & $\approx70\%$  & 0.30                   \\
			\textbf{b}           & $\gamma_0$              & $\approx89\%$  & 0.15                  \\
			\textbf{}            & $\gamma_3$              & $\approx50\%$  & 0.20                   \\
			\textbf{c}           & $\gamma_0$              & $\approx10\%$  & 0.35                  \\
			\textbf{}            & $\gamma_1$              & $\approx90\%$  & 0.03                  \\
			\textbf{d}           & $\gamma_1$              & $\approx23\%$  & 0.10                   \\
			\textbf{}            & $\gamma_4$              & $\approx16\%$  & 0.20                   \\
			\textbf{e}           & $\gamma_0$              & $\approx18\%$  & 0.05                  \\
			\multicolumn{1}{l}{} & $\gamma_1$              & $\approx88\%$  & 0.20                   \\
			\textbf{f}           & $\gamma_0$              & $\approx 7\%$  & 0.05                  \\
			\multicolumn{1}{l}{} & $\gamma_1$              & $\approx62\%$  & 0.01                  \\
			\multicolumn{1}{l}{} & $\gamma_3$              & $\approx26\%$  & 0.03                  \\
			\textbf{g}           & $\gamma_1$              & $\approx64\%$  & 0.03                  \\
			\textbf{}            & $\gamma_3$              & $\approx28\%$  & 0.03                  \\
			\textbf{h}           & $\gamma_0$              & $\approx18\%$  & 0.05                  \\
			& $\gamma_1$              & $\approx88\%$  & 0.20
		\end{tabular}
		
	\end{table}

	\begin{table}[!b]
		\centering
		\caption{	\label{tab2} Table of comparisons of preditictions \emph{vs} ground truths  between traditional adaptive algorithms and our fuzzy algorithms}
		\begin{tabular}{|@{\hspace{0\tabcolsep}}c@{\hspace{0.3\tabcolsep}}|@{\hspace{0.3\tabcolsep}}r@{\hspace{0.3\tabcolsep}}|@{\hspace{0.3\tabcolsep}}c|@{\hspace{0.3\tabcolsep}}|@{\hspace{0.3\tabcolsep}}c|@{\hspace{0.3\tabcolsep}}c|@{\hspace{0.3\tabcolsep}}c|@{\hspace{0.3\tabcolsep}}|}%
			\multicolumn{6}{l}{\textbf{a})Average results for $Th=[0.25,0.45,0.65] $ and $a_1=3$  }\\
			\hline%
			\hline
			
			$Th$&Metric&\emph{Bradley}&$F_{A_4}(Cho.)$&$F_{A_3}(Ham.)$&$F_{A_2}(CF_{1,2})$\\%
			\hline%
			0.25&  $MCC$ &0.65$\pm$0.20&0.65$\pm$0.20&0.65$\pm$0.20&0.32$\pm$0.27\\%
			&    $F_1$    &0.72$\pm$0.18&0.72$\pm$0.18&0.72$\pm$0.18&0.40$\pm$0.26\\%
			&    $SSIM$  &0.69$\pm$0.19&0.69$\pm$0.19&0.69$\pm$0.19&0.63$\pm$0.22\\%
			&  $MSE$     &0.18$\pm$0.15&0.18$\pm$0.15&0.18$\pm$0.15&0.30$\pm$0.24\\%
			& $Acc$  &0.82$\pm$0.15&0.82$\pm$0.15&0.82$\pm$0.15&0.70$\pm$0.24\\%
			& $Prec$ &0.68$\pm$0.26&0.68$\pm$0.26&0.68$\pm$0.26&0.69$\pm$0.30\\%
			&  $Rec$  &0.87$\pm$0.11&0.87$\pm$0.11&0.87$\pm$0.11&0.46$\pm$0.37\\%
			\hline%
			0.45&$MCC$&0.59$\pm$0.24&0.59$\pm$0.24&0.59$\pm$0.24&0.43$\pm$0.28\\%
			&$F_1$  &0.66$\pm$0.21&0.66$\pm$0.21&0.66$\pm$0.21&0.55$\pm$0.24\\%
			&$SSIM$ &0.67$\pm$0.22&0.67$\pm$0.22&0.67$\pm$0.22&0.58$\pm$0.26\\%
			&$MSE$ &0.22$\pm$0.20&0.22$\pm$0.20&0.22$\pm$0.20&0.32$\pm$0.27\\%
			&$Acc$&0.78$\pm$0.20&0.78$\pm$0.20&0.78$\pm$0.20&0.68$\pm$0.27\\%
			&$Prec$&0.70$\pm$0.29&0.70$\pm$0.29&0.70$\pm$0.29&0.61$\pm$0.31\\%
			&$Rec$&0.79$\pm$0.23&0.79$\pm$0.23&0.79$\pm$0.23&0.73$\pm$0.29\\%
			\hline%
			0.65&$MCC$&0.47$\pm$0.27&0.47$\pm$0.27&0.47$\pm$0.27&$\mathbf{0.70\pm 0.17}$\\%
			&$F_1$&0.54$\pm$0.25&0.54$\pm$0.25&0.54$\pm$0.25&$\mathbf{0.76\pm  0.16}$\\%
			&$SSIM$&0.64$\pm$0.24&0.64$\pm$0.24&0.64$\pm$0.24&$\mathbf{0.74 \pm0.17}$\\%
			&$MSE$ &0.27$\pm$0.23&0.27$\pm$0.23&0.27$\pm$0.23&$\mathbf{0.14 \pm 0.11}$\\%
			&$Acc$&0.73$\pm$0.23&0.73$\pm$0.23&0.73$\pm$0.23&$\mathbf{0.86 \pm 0.11}$\\%
			&$Prec$&0.69$\pm$0.31&0.69$\pm$0.31&0.69$\pm$0.31&$\mathbf{0.72 \pm 0.23}$\\%
			&$Rec$&0.68$\pm$0.35&0.68$\pm$0.35&0.68$\pm$0.35&$\mathbf{0.87 \pm 0.10}$\\%
			
			\hline
			\multicolumn{6}{c}{\textbf{b}) Average results with fixed: $a_1=3$}\\
			\hline%
			& &Niblack&Sauvola&\multicolumn{2}{l|}{\multirow{2}{*}{}}  \\%
			\hline%
			&$MCC$&0.48$\pm$0.15&0.48$\pm$0.15&\multicolumn{2}{l|}{} \\%
			&$F_1$&0.60$\pm$0.15&0.60$\pm$0.15&\multicolumn{2}{l|}{} \\%
			&$SSIM$&0.53$\pm$0.16&0.53$\pm$0.16&\multicolumn{2}{l|}{} \\%
			&$MSE$&0.25$\pm$0.09&0.25$\pm$0.09&\multicolumn{2}{l|}{} \\%
			&$Acc$&0.75$\pm$0.09&0.75$\pm$0.09&\multicolumn{2}{l|}{} \\%
			&$Prec$&0.52$\pm$0.20&0.52$\pm$0.20&\multicolumn{2}{l|}{} \\%
			&$Rec$&0.79$\pm$0.11&0.79$\pm$0.11&\multicolumn{2}{l|}{} \\%
			\hline	
		\end{tabular}
		
	\end{table}

	\section{Results and Discussion}
	\label{res}
	
	This Section is organized as follows:
	\textbf{(i)} In Section \ref{comp1}, several analyses on the $toy$ dataset are performed with comparisons between \emph{Bradley} algorithm and our algorithms ($A_2 (CF_{1.2}), A_4 (Choquet)$ and $A_3 (Hamacher)$). \textbf{(ii)} Instead, in Section \ref{comp3}, a whole comparison on the $test$ set with a fixed parametrization between traditional adaptive algorithms and our novel algorithms is described. \textbf{(iii)} Finally, on the same $test$ set, a comparison from the optimal predictions from \emph{DSS-Net} and our novel algorithms is shown in Section \ref{comp2}. For all the setups, the binarized images are tested on the GTs with 7 metrics: Structural Similarity Index ($SSIM$), Mean Square Error ($MSE$), accuracy ($Acc$), precision ($P$), recall ($R$) and the $F_1$ measure ($F_1$) \cite{ntirogiannis2008objective}. In addition, the Matthews Correlation Coefficient metric ($MCC$) \cite{boughorbel2017optimal} is evaluated for the \emph{MSRA-B} binarizations, to deal with the class imbalance problem in real-world images.  The metrics are normalized in the range $[0,1]$, except for MCC and SSIM which are defined in the range $[-1,1]$.
		\begin{table}[t]
		\centering
		\caption{  Pairwise comparisons of our \emph{FLAT} algorithms and \emph{Bradley} algorithm under the same parameter configurations.}
		\begin{tabular}{|l|l@{\hspace{0.5\tabcolsep}}l@{\hspace{-0.05\tabcolsep}}|l@{\hspace{0.5\tabcolsep}}l@{\hspace{-0.05\tabcolsep}}|l@{\hspace{0.5\tabcolsep}}l@{\hspace{-0.05\tabcolsep}}|}
			\hline
			& $g_{FLAT}^*$   & $g_{Bradley}^*$ & $g_{FLAT}^*$   & $g_{Bradley}^*$  &$g_{FLAT}^*$   & $g_{Bradley}^*$ \\
			\hline
			$F_{A_1}$ & 0    &  \textbf{268}  & 0  & \textbf{1101}      & 461  & \textbf{3180} \\
			$F_{A_2}$ & \textbf{850}  &  268  & \textbf{2159} & 1012    & \textbf{3022} & 1162\\
			$F_{A_3}$ & \textbf{260}  &  193  & \textbf{1298} & 677     & \textbf{3107} & 1606\\
			$F_{A_4}$ & 139  &  \textbf{261}  &  460 & \textbf{985}     & \textbf{2670} & 2300 \\
			$SSIM$ & $\geq 0.90$  &     &  $\geq 0.55$  &      & $\geq 0.00$ &  \\
			\hline
		\end{tabular}
		\label{Table_01}
	\end{table}
	\subsection{Comparisons on the Toy dataset}
	\label{comp1}
	\subsubsection{Exahustive analyses} A grid search was carried out on all the possible algorithm parameter configurations to find the \emph{optimal} fuzzy thresholding for the \emph{toy} dataset. A voting schema is suited for comparisons. For what is concerning $w_n$, we tested all the possible windows $n_a \times n_a$ constrained by Eq.\ref{eq8} and $ 1 \leq n_a \leq   \min (n,m) $, where $n,m$ are the dimensions of $I$. 
		Instead, for what is concerning the threshold $Th$, we analyzed all the possible configurations with respect to $w_n$ changing $Th$ increasingly from $0.01$ to $1$ with steps of $0.01$.
		For all the possible parameter configurations of $Th, a_1$ and $a_2$, the binarizations obtained with our algorithms and with the \emph{Bradley} algorithms are divided into three subsets with respect to three specific sensitivity values. Thus, the obtained binarizations are regrouped with respect to $SSIM$ values that are greater or equal to $\theta=[0.90, 0.55, 0.00]$, respectively.  Under the same parameter configurations, the variable $g^*$ indicates the overall number of times when our strategies binarize better than the \emph{Bradley} algorithm and vice versa.
		In particular, in order to obtain a strong pairwise comparison, a counting is made of the times in which the $SSIM$ of the best algorithm is greater then $\theta$ and the algorithm is better than the other. Formally, for each $j=[2,3,4]$, the $i$-th image and the $n$-th parameter configuration, $g^*_{F_{A_j}}$ represents the number of times in which  $SSIM_{F_{A_j}} \geq \theta \land  SSIM_{F_{A_j}} >  SSIM_{Bradley}$ are satisfied. While,  $g^*_{Bradley}$ represents the number of images in which  $SSIM_{Bradley} \geq \theta \land SSIM_{F_{A_j}} <  SSIM_{Bradley}$ are satisfied.	
		These results are shown in Table \ref{Table_01}. For example, in the subset of binarizations with $SSIM \geq 0.9$, the $SSIM$ of $F_{A_2}$ is $g_{FLAT}^* = 850$ times.	As shown in Table \ref{Table_01}, $F_{A_j}$s computed with $A_2$ and $A_3$ are the ones that better binarize the images. Moreover, between the two definitions of fuzzy integrals, our $A_2$ approach is both the best-performing one and that with lower computational complexity than others.

	\begin{table*}[!ht]
		\centering
		\caption{ 	\label{tab3_ott}Table of comparisons of preditictions vs ground truths between our fuzzy algorithms and Bradley at the $optimum$ and the trained DSS-Net \cite{hou2017deeply} }
		\begin{tabular}{|@{\hspace{1.5\tabcolsep}}c@{\hspace{1.5\tabcolsep}}|@{\hspace{1.5\tabcolsep}}c@{\hspace{1.5\tabcolsep}}|@{\hspace{1.5\tabcolsep}}c@{\hspace{1.5\tabcolsep}}|@{\hspace{1.5\tabcolsep}}c@{\hspace{1.5\tabcolsep}}|@{\hspace{1.5\tabcolsep}}c @{\hspace{1.5\tabcolsep}}|@{\hspace{1.5\tabcolsep}}c|@{\hspace{1.5\tabcolsep}}c@{\hspace{1.5\tabcolsep}}|}%
			\hline%
			{-}&$a_1$&\emph{DSS-Net}&\emph{Bradley}&$F_{A_4}$(Choq.)&$F_{A_3}$(Ham.)&$F_{A_2}(CF_{1,2})$ \\%
			\hline%
			$accuracy$&2&0.90$\pm$0.09&0.88$\pm$0.12&0.87$\pm$0.12&0.87$\pm$0.12&$\mathbf{0.94 \pm 0.04}$\\%
			&3&''&0.83$\pm$0.14&0.83$\pm$0.14&0.83$\pm$0.14&$\mathbf{0.91 \pm 0.05}$\\%
			\hline%
			$F_1$&2&0.69$\pm$0.26&0.80$\pm$0.16&0.80$\pm$0.16&0.80$\pm$0.16&$\mathbf{0.89 \pm 0.06}$\\%
			&3&''&0.75$\pm$0.17&0.75$\pm$0.17&0.75$\pm$0.17&$\mathbf{0.83 \pm 0.08}$\\%
			\hline%
			$MCC$&2&0.69$\pm$0.23&0.75$\pm$0.18&0.75$\pm$0.18&0.75$\pm$0.18&$\mathbf{0.86 \pm 0.07}$\\%
			&3&''&0.68$\pm$0.20&0.68$\pm$0.20&0.68$\pm$0.20&$\mathbf{0.78 \pm 0.10}$\\%
			\hline%
			$precision$&2&$\mathbf{0.99 \pm 0.07}$&0.76$\pm$0.22&0.76$\pm$0.22&0.76$\pm$0.23&$ 0.90 \pm 0.07$\\%
			&3&''&0.70$\pm$0.24&0.70$\pm$0.24&0.70$\pm$0.24&$0.84 \pm 0.10  $\\%
			\hline%
			$recall$&2&0.58$\pm$0.26&$\mathbf{0.91\pm0.09}$&$\mathbf{0.91 \pm 0.09}$&$\mathbf{0.91 \pm 0.09}$&$0.88 \pm 0.08$\\%
			&3&''&$\mathbf{0.89 \pm 0.11}$&$\mathbf{0.89 \pm 0.11}$&$\mathbf{0.89 \pm 0.11}$&0.83$\pm$0.10\\%
			\hline%
			$SSIM$&2&$\mathbf{ 0.82 \pm 0.09}$&0.75$\pm$0.18&0.75$\pm$0.18&0.75$\pm$0.18&$\mathbf{ 0.82 \pm 0.09}$\\%
			&3&''&0.69$\pm$0.20&0.69$\pm$0.20&0.69$\pm$0.20&0.79$\pm$0.12\\%
			\hline%
			$MSE$&2&0.10$\pm$0.09&0.12$\pm$0.12&0.13$\pm$0.12&0.13$\pm$0.12&$\mathbf{ 0.06 \pm 0.04}$\\%
			&3&''&0.17$\pm$0.14&0.17$\pm$0.14&0.17$\pm$0.14&$\mathbf{0.09 \pm 0.05}$\\%
			\hline
			\hline
			\multicolumn{7}{|c|}{Average \emph{optimal} threshold values}\\
			\hline
			& & &$Th^*(Brad)$ &$Th^*(F_{A_4})$ &$Th^*(F_{A_3})$&$Th^*(F_{A_2})$\\%
			\hline
			&2&''&0.26$\pm$0.12&0.26$\pm$0.12&0.26$\pm$0.12&0.59$\pm$0.15\\%
			&3&''&0.25$\pm$0.11&0.25$\pm$0.11&0.25$\pm$0.11&0.56$\pm$0.14\\%
			
			\hline   
			
		\end{tabular}
		
	\end{table*}

	\subsubsection{Robustness and sensitivity analyses} An exhaustive analysis on our \emph{toy} dataset with 4 increasing percentages of random additive noise ($+20\%,+30\%,+40\%,+50\%$ of  $\gamma_2$) is provided on the online repository.  In particular, the random noise is added twice, on both the images including the other perturbations ($\gamma_i$ + $\gamma_2, i\neq 2$ ) and on the ground truth images (GT +$\gamma_2$). 
	As it is shown on the online repository, the approach based on $A_2$ is very stable and binarizes with an average $F_1 \geq 0.95$ in the $75\%$ of the cases, and with $0.87 \leq F_1 \leq 0.93$ in the $25\%$ of the cases with the $20\%$ of additive random noise. Instead, considering the $40\%$ of coverage by using $\gamma_2$ perturbation, the $A_2$ methodology binarizes with an average $F_1 \geq 0.95$ in the $62\%$ of the cases, and with $ 0.75 \leq F_1 \leq 0.93$ in the $38\%$ of the cases. While, considering the $A_3$-based methodology with the $40\%$ of $\gamma_2$ coverage, the binarization is maintained up to an average $F_m \geq 0.89$ in the $56\%$ of the cases, and with $0.53  \leq F_1 \leq 0.78$ in the rest of the cases.
	
	\subsubsection{Qualitative and comparative analyses}
	In Table \href{https://github.com/lodeguns/FuzzyAdaptiveBinarization}{S-1} on the online repository, our algorithm binarizations (respectively $I_{b}$, for $b=A_2, A_3, A_4 $) and the \emph{Bradley} algorithm binarizations ($I_B$) are compared with the same parameter configurations ($n_a$ and $t$).  In  Table \href{https://github.com/lodeguns/FuzzyAdaptiveBinarization}{S-1} the best values are indicated with asterisks and in bold. The toy images present different types of perturbations, that differently affects an accurate binarization. In Table \ref{per_table} the percentages of applied perturbations are indicated.   As it is shown in Table \href{https://github.com/lodeguns/FuzzyAdaptiveBinarization}{S-1}, the comparative analysis indicates that the \emph{Bradley} algorithm seems to be stable only with high-low contrast variations ($\gamma_0$) and spatial variation in lightning ($\gamma_1 $) (see also Figure \ref{fig_01} - Image \textbf{c}). While, with our fuzzy algorithms, the binarization is more stable with all the types of perturbations considered.  This is proved also with a visual example in Figure \ref{fig_01}, where the binarization of Image \textbf{a} is shown. In fact, in this case, Image \textbf{a} presents a high percentage of $\gamma_0$ and $\gamma_3$ (see Table \ref{per_table}). The latter perturbation represents motifs (recurrent patterns) of structured noise which are very difficult to threshold. For other visual comparisons please visit our online repository. The \emph{FLAT} methodology based on $A_1$ does never reach good results, however, its analyses are shown in the online repository.

	\subsection{Comparisons with traditional algorithms}
	\label{comp3}

\begin{figure*}[!htbp]
		\centering
		\begin{tabular}{@{\hspace{-0.2\tabcolsep}}c
				@{\hspace{-0.2\tabcolsep}}c
				@{\hspace{-0.2\tabcolsep}}c
				@{\hspace{-0.2\tabcolsep}}c
				@{\hspace{-0.2\tabcolsep}}c
				@{\hspace{-0.2\tabcolsep}}c
				@{\hspace{-0.2\tabcolsep}}c
				@{\hspace{-0.2\tabcolsep}}c
				@{\hspace{-0.2\tabcolsep}}c
				@{\hspace{-0.2\tabcolsep}}c
				@{\hspace{-0.2\tabcolsep}}  }
			\\
			\frame{\includegraphics[width=18.5mm]{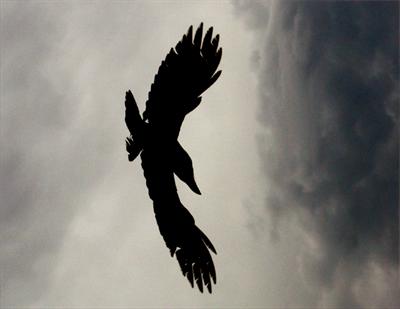}}&
			\frame{\includegraphics[width=18.5mm]{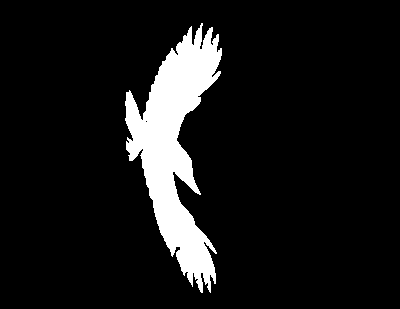}}&
			\frame{\includegraphics[width=18.5mm]{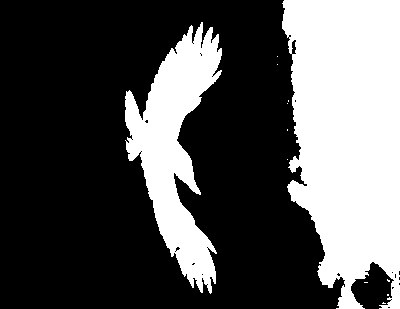}}&
			\frame{\includegraphics[width=18.5mm]{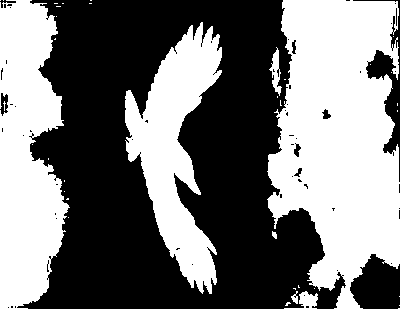}}&
			\frame{\includegraphics[width=18.5mm]{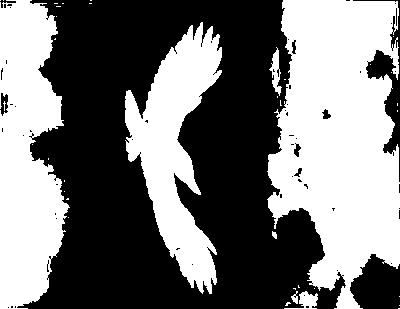}}&   \frame{\includegraphics[width=18.5mm]{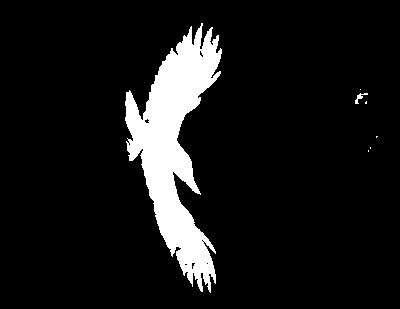}}&
			\frame{\includegraphics[width=18.5mm]{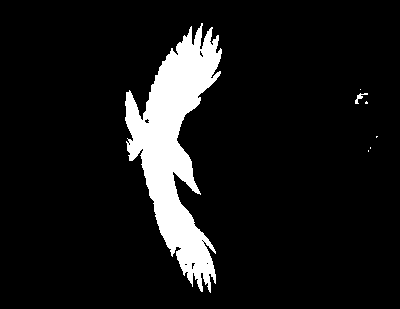}}&   \frame{\includegraphics[width=18.5mm]{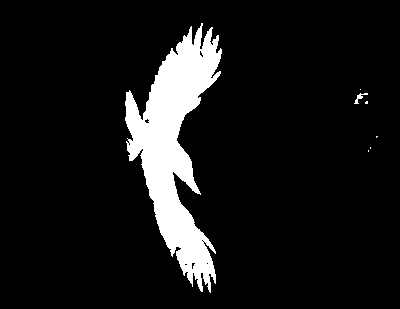}}&
			\frame{\includegraphics[width=18.5mm]{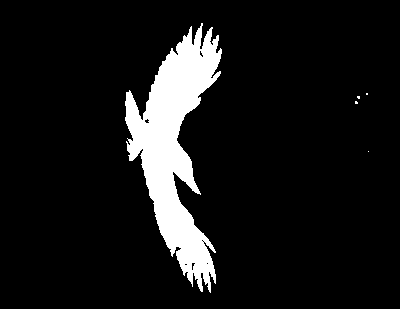}}&   \frame{\includegraphics[width=18.5mm]{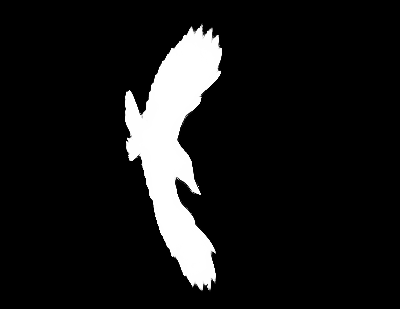}}\\
			
			\frame{\includegraphics[width=18.5mm]{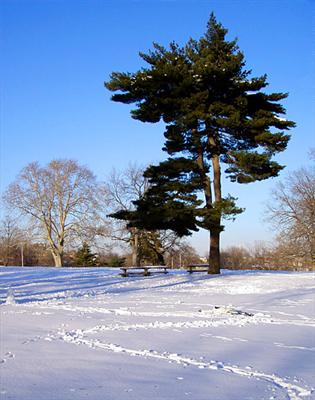}}&
			\frame{\includegraphics[width=18.5mm]{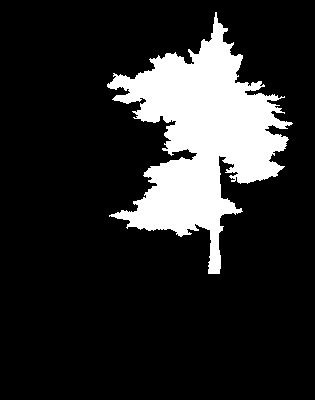}}&
			\frame{\includegraphics[width=18.5mm]{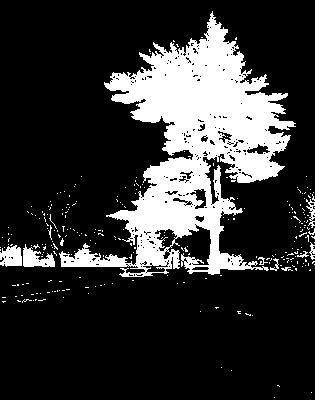}}&
			\frame{\includegraphics[width=18.5mm]{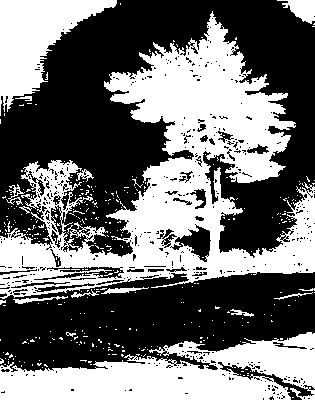}}&
			\frame{\includegraphics[width=18.5mm]{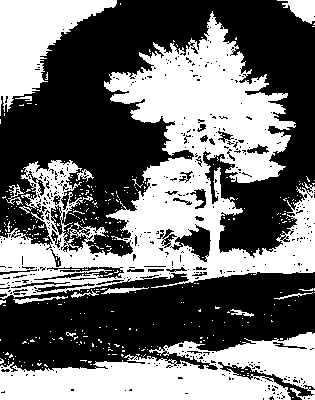}}&   \frame{\includegraphics[width=18.5mm]{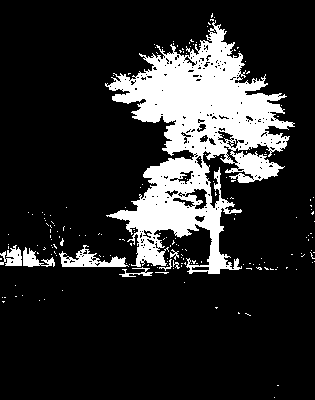}}&
			\frame{\includegraphics[width=18.5mm]{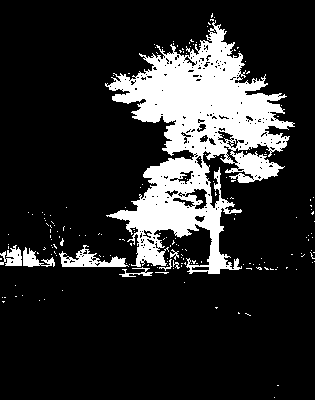}}&   \frame{\includegraphics[width=18.5mm]{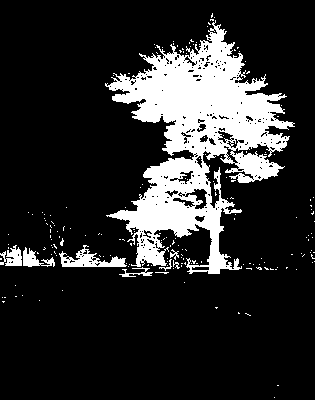}}&
			\frame{\includegraphics[width=18.5mm]{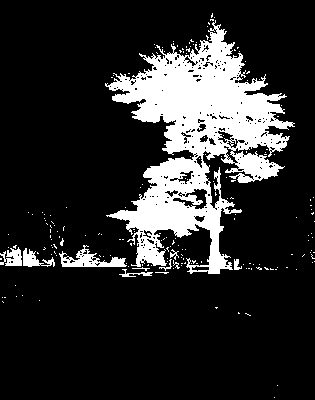}}&   \frame{\includegraphics[width=18.5mm]{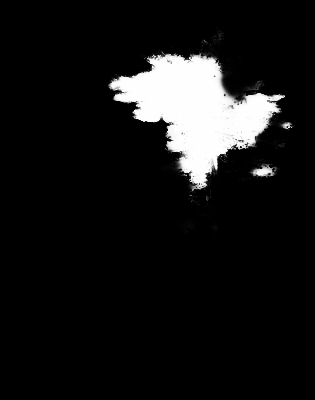}}\\
			
			\frame{\includegraphics[width=18.5mm]{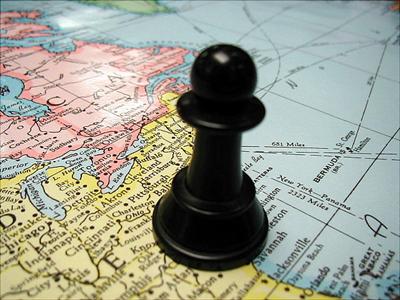}}&
			\frame{\includegraphics[width=18.5mm]{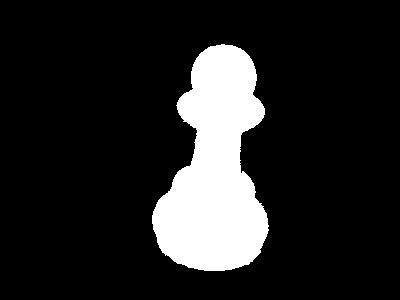}}&
			\frame{\includegraphics[width=18.5mm]{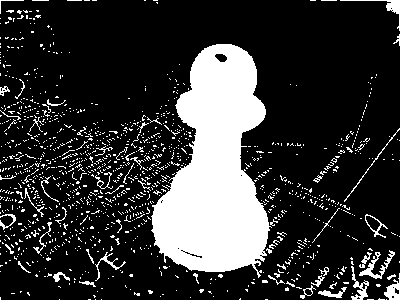}}&
			\frame{\includegraphics[width=18.5mm]{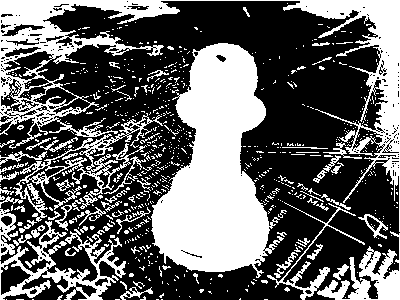}}&
			\frame{\includegraphics[width=18.5mm]{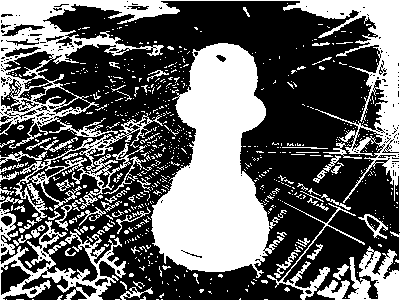}}&   \frame{\includegraphics[width=18.5mm]{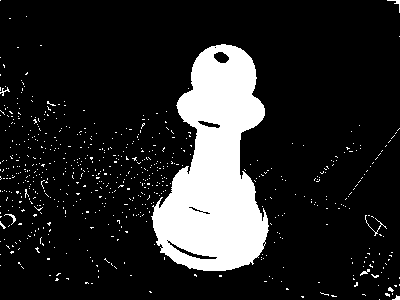}}&
			\frame{\includegraphics[width=18.5mm]{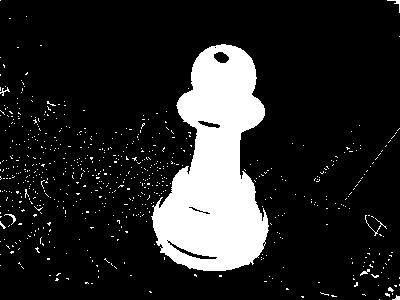}}&   \frame{\includegraphics[width=18.5mm]{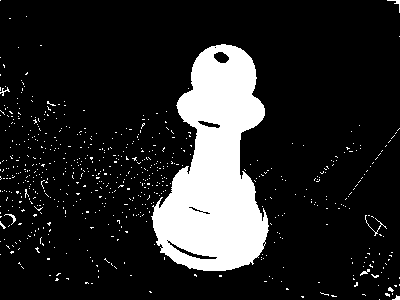}}&
			\frame{\includegraphics[width=18.5mm]{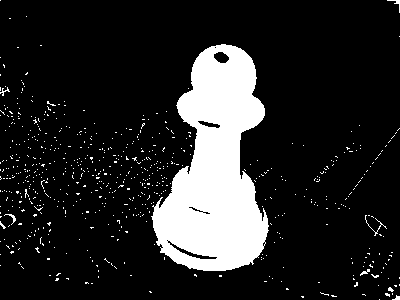}}&
			\frame{\includegraphics[width=18.5mm]{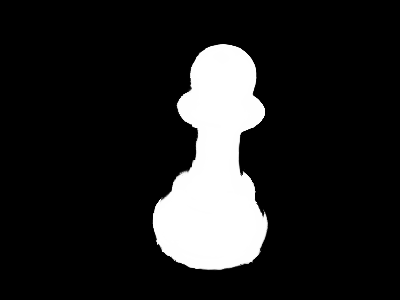}} \\
			
			\frame{\includegraphics[width=18.5mm]{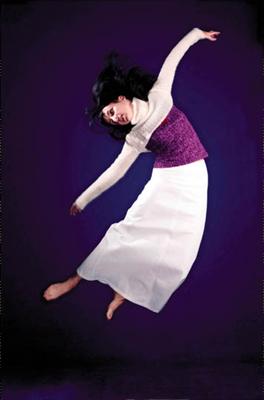}}&
			\frame{\includegraphics[width=18.5mm]{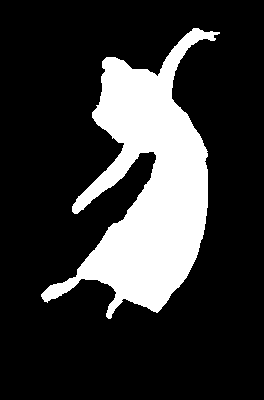}}&
			\frame{\includegraphics[width=18.5mm]{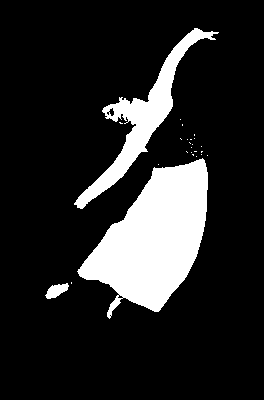}}&
			\frame{\includegraphics[width=18.5mm]{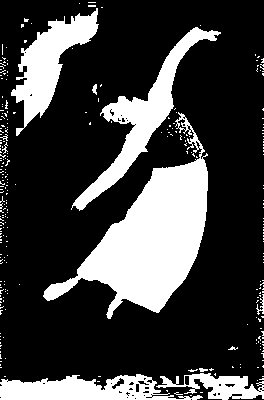}}&
			\frame{\includegraphics[width=18.5mm]{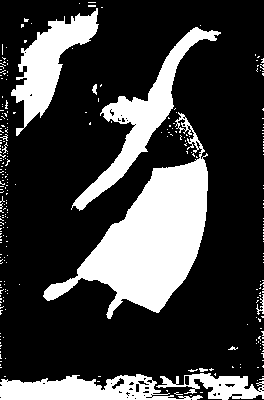}}&   \frame{\includegraphics[width=18.5mm]{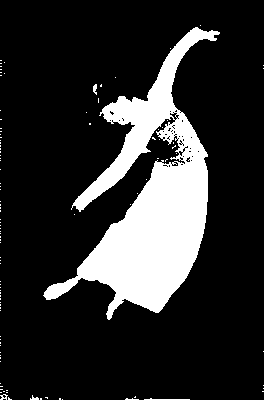}}&
			\frame{\includegraphics[width=18.5mm]{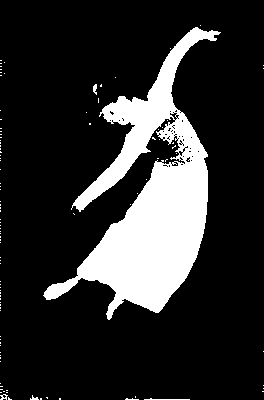}}&   \frame{\includegraphics[width=18.5mm]{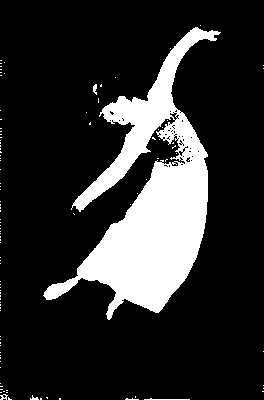}}&
			\frame{\includegraphics[width=18.5mm]{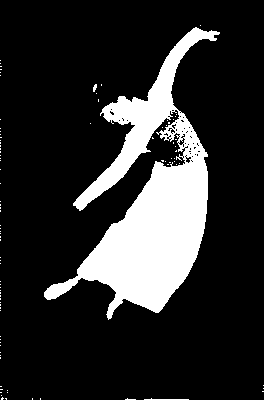}}&   \frame{\includegraphics[width=18.5mm]{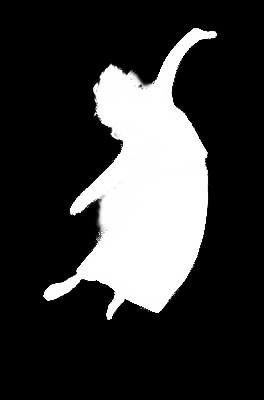}}
			
			\\
		\emph{Original} & \emph{Ground truth}    & $Otsu$  & \emph{Niblack} & \emph{Sauvola} & \emph{Bradley} & $A_4$  &  $A_3$  & $A_2$  & \emph{DSS-Net}
			
		\end{tabular}
		\caption{ 	\label{fig001}
			In  Figure \ref{fig001} a visual comparison of binary maps produced by the proposed algorithms ($A_4$ - Choquet, $A_3$ - Hamacher,  $A_2$ - $CF_{1,2}$ generalization), the global thresholding algorithm (\emph{Otsu}), the local thresholding algorithms (\emph{Niblack}, \emph{Sauvola} and \emph{Bradley}) is shown. The produced binarisations look similar and generally better than the traditional ones. In particular, the adaptive behaviour of $A_4$, $A_3$ and $A_2$ turn out to be more relevant in reliable documents, and on images with shadows.  }
	\end{figure*}

		Our fuzzy algorithms based on $A_2$, $A_3$ and $A_4$ functions were tested on the $test$ set of 2413 images (see also sub-section \ref{datasets}) with respect to adaptive methods of \emph{Sauvola} \cite{sauvola2000adaptive}, \emph{Niblack}\cite{niblack1986introduction} and \emph{Bradley and Roth}.
		As it is shown in Table \ref{tab2} \textbf{(a)}, for what concerns our algorithms and the algorithm of \emph{Bradley et al.}\cite{bradley2007adaptive}, three different threshold levels ($Th=[0.25, 0.45, 0.75]$) and two fixed window parameters $a_1=3, a_2=1$ were chosen. On the other hand, for \emph{Niblack} and \emph{Sauvola} ( Table \ref{tab2} \textbf{(b)}) the results are computed considering only the same fixed window parameters. In this case, it is impossible to fix a threshold because these algorithms compute their adaptive threshold value basing their binarizations on the mean and standard deviation of the window centered on the pixel to binarize.
		Furthermore, it is important to underline that two other parameters have been fixed, in order to make the comparisons as balanced as possible. In particular, the parameter \emph{K} for \emph{Niblack} is set to 0, because in such a way, exhibits a generalized behavior like the \emph{Bradley} algorithm. While, as suggested  by \emph{Sauvola et al.}\cite{sauvola2000adaptive}, the values of \emph{K} and  \emph{R} are set to 0.2 and 128, respectively.   As it is shown in Table \ref{tab2}, our algorithms outperform the binarizations obtained with \emph{niblack} and \emph{Sauvola}, and, in particular, our methodology based on $A_2 (CF_{1,2})$ turns out to be the best performing one, with a threshold fixed to $0.65$. In  Figure \ref{fig001}, a visual comparison of binary maps produced by the proposed algorithms is shown. By looking at the obtained binarizations, also with the \emph{DSS-Net} predictions (see also sub-section \ref{comp2}), our proposed algorithms turn out to be more reliable in printed documents, and on images with shadows.
		From a first analysis, even if $A_2 (CF_{1,2})$ seems to be the one that performs better, there was no big difference for the threshold values at 0.25 and 0.45 between our algorithms $A_4 (Choquet)$, $A_3 (Hamacher)$ and \emph{Bradley}'s. Moreover, in this case, the thresholds were chosen empirically; on the other hand, as we will show in the next paragraph, at the $optimum$, the quality of our binarizations is better than those obtained with the \emph{Bradley} algorithm.

	\subsection{Comparisons with DSS-Net and Bradley at the optimum}
	\label{comp2}

	At best of our knowledge, Deep Learning models have not been used in image thresholding. Few attempts have been done so far for solving similar tasks as RED-Net (Residual Encoder-Decoder Network - U-Net \cite{calvo2019selectional}) for hand-written document binarization and Le-Net5  \cite{kung2018efficient,calvo2017pixel} (a traditional CNN based on the model of \cite{lecun1998gradient}) for musical document binarization.
	In this study we chose DSS-Net \cite{hou2017deeply}, that is a CNN trained for saliency on real world images, which we used in our experimental set-up. As far as we know, \emph{DSS}-Net seems to be the best comparable model concerning our adaptive algorithms, because it retains a strong generalization power, deriving from the use of a very extensive data set on binarizable images.
	For the comparisons between \emph{DSS}-Net, \emph{Bradley} and our fuzzy algorithms, only $280$ images with GTs were selected on the $test$ set (see Section \ref{datasets}).
	In particular, the images were selected considering an \emph{Otsu} $F_1$ measure greater or equal to $0.8$ to ensure a reliable level of thresholdability. Moreover, only the images that are more difficult to be binarized have been selected with a manual control. In fact, they are complex in terms of shading and lighting, relative positions of objects and variable size of objects in the background and foreground. The subset of these thresholdable images with their predicted binary masks is available on the online repository .
	For each image, the binarization of \emph{Bradley} and our methodologies are computed at the $optimum$, selecting only the best results. In particular, the search of the $optimum$ is obtained changing $Th$ increasingly from $0.01$ to $1$  with steps of $0.01$.
	In Table \ref{tab3_ott}, the average results of these comparisons are shown for several metrics.
	In particular, the table shows that $ F_ {A_2} $  reaches an MCC $\approx 0.86 \pm 0.07$  showing the algorithm ability to manage binarizations with a very different ratio between the pixels classified as background and foreground, dealing correctly with true and false positives and negatives. For what is concerning the \emph{precision},  \emph{DSS-Net} and $ F_ {A_2} $ show a better ability to recognize false positives. Looking at the \emph{recall}, \emph{Bradley}, and our $F_{A_4}$ and $F_{A_3}$ are more accurate in the detection of false negatives. However, the best accurate $F_1$, which is more stable on extreme values, and $accuracy$, according to the $MCC$, is obtained with $ F_ {A_2}( F_1 = \approx 0.8, accuracy = \approx 0.9) $. The similarity between predictions/binarizations and GTs are evaluated considering, also the presence of noise, with the $SSIM$. In this case, \emph{DSS}-Net reaches the same performance of $ F_ {A_2} $ with $a_1=2$ and $ a_2=1$.  For what is concerning $MSE$, $F_ {A_2}$ outperform all the other algorithms with the different $a_2$ configurations.  In conclusion, similarly to the experiments on the $toy$ dataset (see section \ref{comp1} and to the experiments on the dataset of 2413 images (see section \ref{comp3}), the Choquet ($ A_4 $) and Hamacher ($ A_3 $) methodologies, show equal and slightly lower performances than those of $ A_2 $ and \emph{DSS}-Net by varying the $a_1$ parameter and fixing $a_2=1$. Furthermore, as it is described above, the \emph{FLAT} methodology based on $CF_{1_2}$ seems to perform much better than the other fuzzy algorithms and \emph{DSS}-Net predictions.
	Moreover, Google Colab \cite{carneiro2018performance}, a benchmark of 10 images with a fixed size of $200 \times 200$ pixels, was selected for evaluating the binarization time of our algorithms.  In detail, our fuzzy algorithms reach $\approx 1100fps$. For what is concerning DSS-Net, authors declare a prediction time of $\approx 750fps$ on images with a variable size. Moreover, it is important to underline that the search time of the optimal threshold is a limitation for our algorithms. Therefore, the search time can greatly reduce the number of frames per second in binarization. The search range of the optimum could be restricted, beacause as it is shown on Table \ref{tab3_ott}, on an extended dataset of images, the average of the $optimal$ threshold values seem to settle on certain values, with a very low standard deviation.

	\section{Conclusion}
	 Three new models for adaptive binarization, based on the optimized calculation of generalized fuzzy integral images, were introduced. The algorithm optimizations were obtained through a novel modification of the summed area table algorithm which, in particular, is suited for fuzzy integrals. It has been shown that, compared to traditional methods and a state of the art neural network, our adaptive methods have improved the accuracy of binarization  without additional computational complexity. In particular, according the the $MCC$ and $F_1$ metrics, one of our proposed algorithms ($CF_{1,2}$) reaches $F_1 \approx 0.86$ with  standard deviation of $\approx 0.07$ and $MCC \approx 0.82$ and a standard deviation of $0.04$. Fuzzy thresholding algorithms turn out to be very stable for a correct thresholding of real-world images which are highly perturbed by different lighting conditions,  background/foreground size imbalance and several color contrast conditions. Due to the impressive time performances ($1100fps$), these new thresholding algorithms could be embedded in deep learning models obtaining a better accuracy and speeding up the network convergence. In conclusion, the results obtained in this paper, both from a theoretical and applied points of view, are really promising. We expect that these novel methodologies will lead to new research opportunities in real time binarization and image processing.

	\appendices

	\ifCLASSOPTIONcompsoc
	\section*{Acknowledgments}
	\else
	\section*{Acknowledgment}
	\fi
	
	This work is supported by  Programma Operativo Nazionale FSE-FESR “Ricerca Innovazione 2014-2020”, Asse I “Capitale Umano”, Azione I.1 “Dottorati Innovativi con caratterizzazione industriale”, DOT1728107 - MIUR (Italy) and VEGA 1/0614/18 and VEGA 1/0545/20 and TIN2016-
	77356-P (AEI/FEDER,UE). B. de la Osa, H. Bustince and J. Fernandez were also supported by project
	PC093-094 TFIPDL of the Government of Navarra.

	\ifCLASSOPTIONcaptionsoff
	\newpage
	\fi
	
	\bibliographystyle{ieeetr}
	\bibliography{IEEEabrv,./bib/paper}

\begin{thebibliography}{10}

\bibitem{aich2018}
S.~Aich, W.~van~der Kamp, and I.~Stavness, ``Semantic binary segmentation using
  convolutional networks without decoders,'' in {\em The IEEE Conference on
  Computer Vision and Pattern Recognition (CVPR) Workshops}, June 2018.

\bibitem{cheremkhin2019comparative}
P.~A. Cheremkhin and E.~A. Kurbatova, ``Comparative appraisal of global and
  local thresholding methods for binarisation of off-axis digital holograms,''
  {\em Optics and Lasers in Engineering}, vol.~115, pp.~119--130, 2019.

\bibitem{kalaiselvi2017comparative}
T.~Kalaiselvi, P.~Nagaraja, and V.~Indhu, ``A comparative study on thresholding
  techniques for gray image binarization,'' {\em Int. J. of Advanced Research
  in Computer Science}, vol.~8, 2017.

\bibitem{roy2014adaptive}
P.~Roy, S.~Dutta, N.~Dey, G.~Dey, S.~Chakraborty, and R.~Ray, ``Adaptive
  thresholding: a comparative study,'' in {\em 2014 International conference on
  control, Instrumentation, communication and Computational Technologies
  (ICCICCT)}, pp.~1182--1186, IEEE, 2014.

\bibitem{achanta2009frequency}
R.~Achanta, S.~Hemami, F.~Estrada, and S.~Susstrunk, ``Frequency-tuned salient
  region detection,'' in {\em 2009 IEEE conference on computer vision and
  pattern recognition}, pp.~1597--1604, IEEE, 2009.

\bibitem{dias2018using}
C.~A. Dias, J.~C. Bueno, E.~N. Borges, S.~S. Botelho, G.~P. Dimuro, G.~Lucca,
  J.~Fernand{\'e}z, H.~Bustince, and P.~L.~J. Drews~Jr., ``Using the {C}hoquet
  integral in the pooling layer in deep learning networks,'' in {\em North
  American Fuzzy Information Processing Society Annual Conference},
  pp.~144--154, Springer, 2018.

\bibitem{dai2015convolutional}
J.~Dai, K.~He, and J.~Sun, ``Convolutional feature masking for joint object and
  stuff segmentation,'' in {\em Proceedings of the IEEE Conference on Computer
  Vision and Pattern Recognition}, pp.~3992--4000, 2015.

\bibitem{he2019deepotsu}
S.~He and L.~Schomaker, ``Deepotsu: Document enhancement and binarization using
  iterative deep learning,'' {\em Pattern Recognition}, vol.~91, pp.~379--390,
  2019.

\bibitem{fan2019road}
R.~Fan, M.~J. Bocus, Y.~Zhu, J.~Jiao, L.~Wang, F.~Ma, S.~Cheng, and M.~Liu,
  ``Road crack detection using deep convolutional neural network and adaptive
  thresholding,'' {\em arXiv preprint arXiv:1904.08582}, 2019.

\bibitem{yan2018weight}
X.~Yan, L.~G. Jeub, A.~Flammini, F.~Radicchi, and S.~Fortunato, ``Weight
  thresholding on complex networks,'' {\em Physical Review E}, vol.~98, no.~4,
  p.~042304, 2018.

\bibitem{gross2019analytical}
T.~J. Gross, M.~Bessani, W.~D. Junior, R.~B. Ara{\'u}jo, F.~A.~C. Vale, and
  C.~D. Maciel, ``An analytical threshold for combining bayesian networks,''
  {\em Knowledge-Based Systems}, vol.~175, pp.~36--49, 2019.

\bibitem{bardozzo2018study}
F.~Bardozzo, P.~Li{\'o}, and R.~Tagliaferri, ``A study on multi-omic
  oscillations in escherichia coli metabolic networks,'' {\em BMC
  bioinformatics}, vol.~19, no.~7, p.~194, 2018.

\bibitem{sauvola2000adaptive}
J.~Sauvola and M.~Pietik{\"a}inen, ``Adaptive document image binarization,''
  {\em Pattern recognition}, vol.~33, no.~2, pp.~225--236, 2000.

\bibitem{bradley2007adaptive}
D.~Bradley and G.~Roth, ``Adaptive thresholding using the integral image,''
  {\em Journal of graphics tools}, vol.~12, no.~2, pp.~13--21, 2007.

\bibitem{goyal2008compressive}
V.~K. Goyal, A.~K. Fletcher, and S.~Rangan, ``Compressive sampling and lossy
  compression,'' {\em IEEE Signal Processing Magazine}, vol.~25, no.~2,
  pp.~48--56, 2008.

\bibitem{wang2018automatic}
Z.~Wang, X.~Huang, and Z.~Cheng, ``Automatic spot identification method for
  high throughput surface plasmon resonance imaging analysis,'' {\em
  Biosensors}, vol.~8, no.~3, p.~85, 2018.

\bibitem{hudaib2016new}
A.~A. Hudaib, H.~N. Fakhouri, and R.~Ghnemat, ``New methodology for microarray
  spot segmentation and gene expression analysis,'' {\em Scientific Research
  and Essays}, vol.~11, no.~12, pp.~126--134, 2016.

\bibitem{el2008fuzzy}
F.~El~Baf, T.~Bouwmans, and B.~Vachon, ``Fuzzy integral for moving object
  detection,'' in {\em 2008 IEEE International Conference on Fuzzy Systems
  (IEEE World Congress on Computational Intelligence)}, pp.~1729--1736, IEEE,
  2008.

\bibitem{boegel2015fully}
M.~Boegel, P.~Hoelter, T.~Redel, A.~Maier, J.~Hornegger, and A.~Doerfler, ``A
  fully-automatic locally adaptive thresholding algorithm for blood vessel
  segmentation in 3d digital subtraction angiography,'' in {\em 2015 37th
  Annual International Conference of the IEEE Engineering in Medicine and
  Biology Society (EMBC)}, pp.~2006--2009, IEEE, 2015.

\bibitem{ciaparrone2019deep}
G.~Ciaparrone, F.~L. S{\'a}nchez, S.~Tabik, L.~Troiano, R.~Tagliaferri, and
  F.~Herrera, ``Deep learning in video multi-object tracking: A survey,'' {\em
  Neurocomputing}, 2019.

\bibitem{zemmour2019automatic}
E.~Zemmour, P.~Kurtser, and Y.~Edan, ``Automatic parameter tuning for adaptive
  thresholding in fruit detection,'' {\em Sensors}, vol.~19, no.~9, p.~2130,
  2019.

\bibitem{wellner1993adaptive}
P.~D. Wellner, ``Adaptive thresholding for the digitaldesk,'' {\em Xerox,
  EPC1993-110}, pp.~1--19, 1993.

\bibitem{niblack1986introduction}
W.~Niblack, ``An introduction to digital image processing, 115--116
  prentice-hall,'' {\em Englewood Cliffs, New Jersey}, 1986.

\bibitem{debayle2009general}
J.~Debayle and J.-C. Pinoli, ``General adaptive neighborhood {C}hoquet image
  filtering,'' {\em Journal of Mathematical Imaging and Vision}, vol.~35,
  no.~3, pp.~173--185, 2009.

\bibitem{wu2015handwritten}
J.~Wu, F.~Da, C.~Wang, and S.~Gai, ``Handwritten character recognition based on
  weighted integral image and probability model,'' in {\em International
  Conference on Image and Graphics}, pp.~347--360, Springer, 2015.

\bibitem{kasagi2014parallel}
A.~Kasagi, K.~Nakano, and Y.~Ito, ``Parallel algorithms for the summed area
  table on the asynchronous hierarchical memory machine, with gpu
  implementations,'' in {\em 2014 43rd International Conference on Parallel
  Processing}, pp.~251--260, IEEE, 2014.

\bibitem{grabisch2009aggregation}
M.~Grabisch, J.-L. Marichal, R.~Mesiar, and E.~Pap, {\em Aggregation
  functions}, vol.~127.
\newblock Cambridge University Press, 2009.

\bibitem{horanska2018generalization}
L.~Horansk{\'a} and A.~{\v{S}}ipo{\v{s}}ov{\'a}, ``A generalization of the
  discrete {C}hoquet and {S}ugeno integrals based on a fusion function,'' {\em
  Information Sciences}, vol.~451, pp.~83--99, 2018.

\bibitem{crow1984summed}
F.~C. Crow, ``Summed-area tables for texture mapping,'' in {\em ACM SIGGRAPH
  computer graphics}, vol.~18, pp.~207--212, ACM, 1984.

\bibitem{el2008foreground}
F.~El~Baf, T.~Bouwmans, and B.~Vachon, ``Foreground detection using the
  {C}hoquet integral,'' in {\em 2008 Ninth International Workshop on Image
  Analysis for Multimedia Interactive Services}, pp.~187--190, IEEE, 2008.

\bibitem{barreto2018fuzzy}
G.~A. Barreto and R.~Coelho, {\em Fuzzy Information Processing: 37th Conference
  of the North American Fuzzy Information Processing Society, NAFIPS 2018,
  Fortaleza, Brazil, July 4-6, 2018, Proceedings}, vol.~831.
\newblock Springer, 2018.

\bibitem{ntirogiannis2008objective}
K.~Ntirogiannis, B.~Gatos, and I.~Pratikakis, ``An objective evaluation
  methodology for document image binarization techniques,'' in {\em 2008 The
  Eighth IAPR International Workshop on Document Analysis Systems},
  pp.~217--224, IEEE, 2008.

\bibitem{jiang2013salient}
H.~Jiang, J.~Wang, Z.~Yuan, Y.~Wu, N.~Zheng, and S.~Li, ``Salient object
  detection: A discriminative regional feature integration approach,'' in {\em
  Proceedings of the IEEE conference on computer vision and pattern
  recognition}, pp.~2083--2090, 2013.

\bibitem{otsu1979threshold}
N.~Otsu, ``A threshold selection method from gray-level histograms,'' {\em IEEE
  transactions on systems, man, and cybernetics}, vol.~9, no.~1, pp.~62--66,
  1979.

\bibitem{hou2017deeply}
Q.~Hou, M.-M. Cheng, X.~Hu, A.~Borji, Z.~Tu, and P.~H. Torr, ``Deeply
  supervised salient object detection with short connections,'' in {\em
  Proceedings of the IEEE Conference on Computer Vision and Pattern
  Recognition}, pp.~3203--3212, 2017.

\bibitem{boughorbel2017optimal}
S.~Boughorbel, F.~Jarray, and M.~El-Anbari, ``Optimal classifier for imbalanced
  data using matthews correlation coefficient metric,'' {\em PloS one},
  vol.~12, no.~6, 2017.

\bibitem{calvo2019selectional}
J.~Calvo-Zaragoza and A.-J. Gallego, ``A selectional auto-encoder approach for
  document image binarization,'' {\em Pattern Recognition}, vol.~86,
  pp.~37--47, 2019.

\bibitem{kung2018efficient}
J.~Kung, D.~Zhang, G.~Van~der Wal, S.~Chai, and S.~Mukhopadhyay, ``Efficient
  object detection using embedded binarized neural networks,'' {\em Journal of
  Signal Processing Systems}, vol.~90, no.~6, pp.~877--890, 2018.

\bibitem{calvo2017pixel}
J.~Calvo-Zaragoza, G.~Vigliensoni, and I.~Fujinaga, ``Pixel-wise binarization
  of musical documents with convolutional neural networks,'' in {\em 2017
  Fifteenth IAPR International Conference on Machine Vision Applications
  (MVA)}, pp.~362--365, IEEE, 2017.

\bibitem{lecun1998gradient}
Y.~LeCun, L.~Bottou, Y.~Bengio, P.~Haffner, {\em et~al.}, ``Gradient-based
  learning applied to document recognition,'' {\em Proceedings of the IEEE},
  vol.~86, no.~11, pp.~2278--2324, 1998.

\bibitem{carneiro2018performance}
T.~Carneiro, R.~V.~M. Da~N{\'o}brega, T.~Nepomuceno, G.-B. Bian, V.~H.~C.
  De~Albuquerque, and P.~P. Reboucas~Filho, ``Performance analysis of google
  colaboratory as a tool for accelerating deep learning applications,'' {\em
  IEEE Access}, vol.~6, pp.~61677--61685, 2018.

\end{thebibliography}


@article{sauvola2000adaptive,
  title={Adaptive document image binarization},
  author={Sauvola, Jaakko and Pietik{\"a}inen, Matti},
  journal={Pattern recognition},
  volume={33},
  number={2},
  pages={225--236},
  year={2000},
  publisher={Elsevier}
}

	\begin{IEEEbiography}[{\includegraphics[width=1.1in,height=1.25in,clip,keepaspectratio]{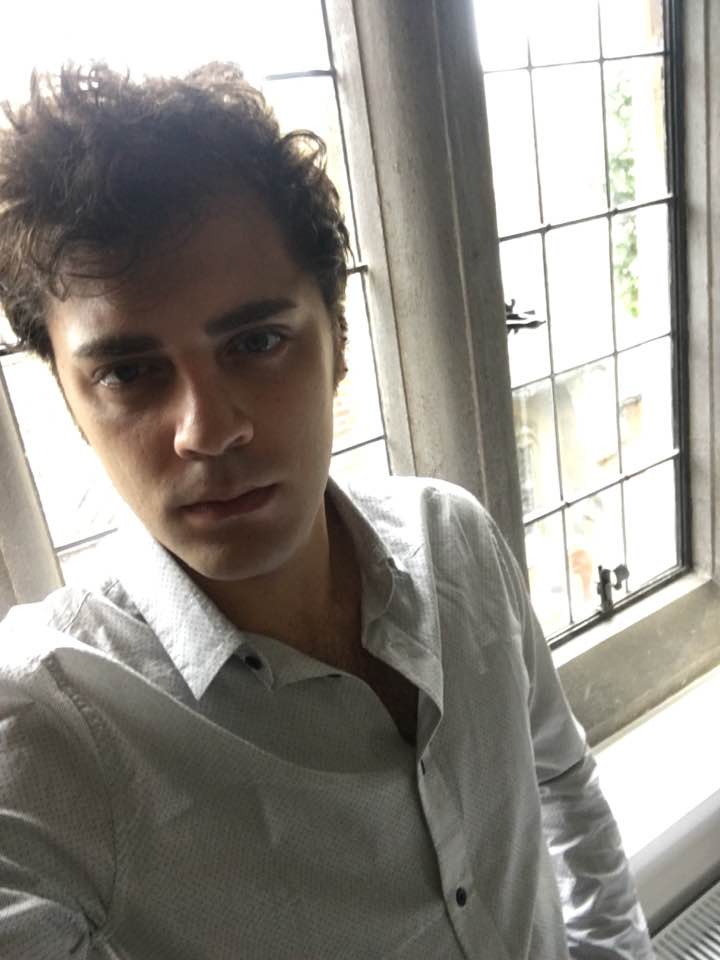}}]{Francesco Bardozzo} received both a B.Sc and M.Sc degree with honors in artificial and computational intelligence from the Faculty of Computer Science - University of Salerno (IT). He is currently Ph.D. student of the DISA-MIS (Department of Business Sciences, Management and Innovation Systems) of the University of Salerno. His research interests include Artificial and Computational Intelligence, Machine Learning, Deep Learning and Computational Biology with special stress in multi-omic oscillations, soft tissue reconstruction and neuroimaging.
	\end{IEEEbiography}
	
	\begin{IEEEbiography}[{\includegraphics[width=1in,height=1.25in,clip,keepaspectratio]{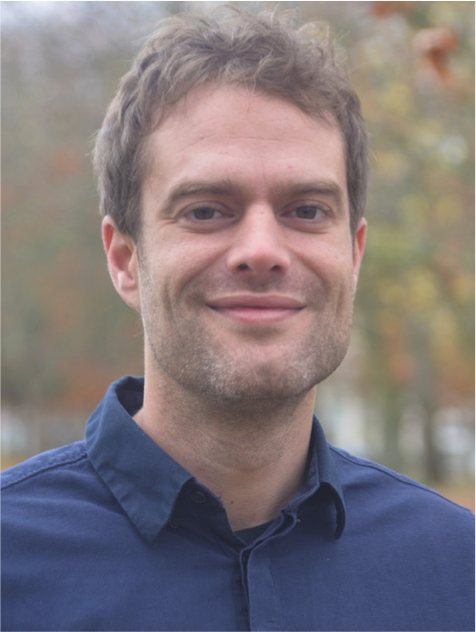}}]{Borja De La Osa} received a B.Sc and M.Sc degree in Industrial Engineering from the Public University of Navarre in 2012. He worked as a Quality Assurance Analyst in the automotive industry for 7 years. He received the degree of Expert in Data Science and Big Data for Business Intelligence from the Public University of Navarre in 2019. He is currently an Associate Professor and Ph.D. student in the Department of Statistics, Computer Science and Mathematics of the Public University of Navarre. His research interests include fuzzy techniques for image processing, unsupervised learning and reinforcement learning.
	\end{IEEEbiography}
	
	\begin{IEEEbiography}[{\includegraphics[width=1in,height=1.25in,clip,keepaspectratio]{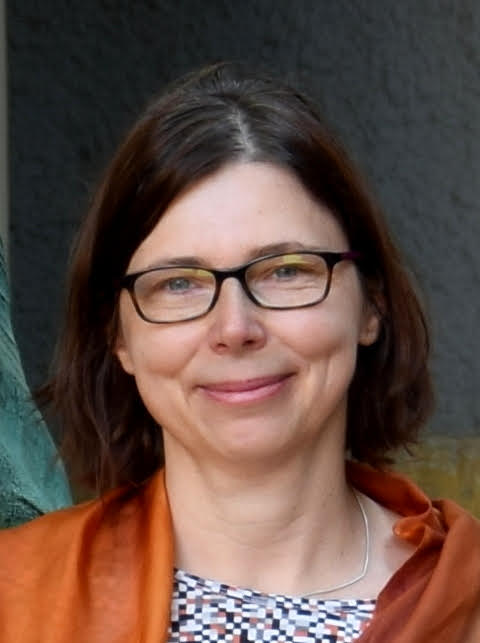}}]{{\mL}ubom\'ira~Horansk\'a}
		received the Graduate degree in mathematics,  and the Ph.D. degree in geometry and topology,  from the Faculty of Mathematics and Physics,  Comenius University, Bratislava, Slovakia, in 1993 and 2001, respectively. Since 1993, she is with the  Institute of Information Engineering, Automation and Mathematics, Faculty of Chemical and Food Technology, Slovak University of Technology in Bratislava,  Slovakia. Her research interests include  uncertainty modeling, aggregation functions, with a special stress to copulas, measures and integrals and algebraic and differetial topology. \end{IEEEbiography}
	
	\begin{IEEEbiography}[{\includegraphics[width=1in,height=1.25in,clip,keepaspectratio]{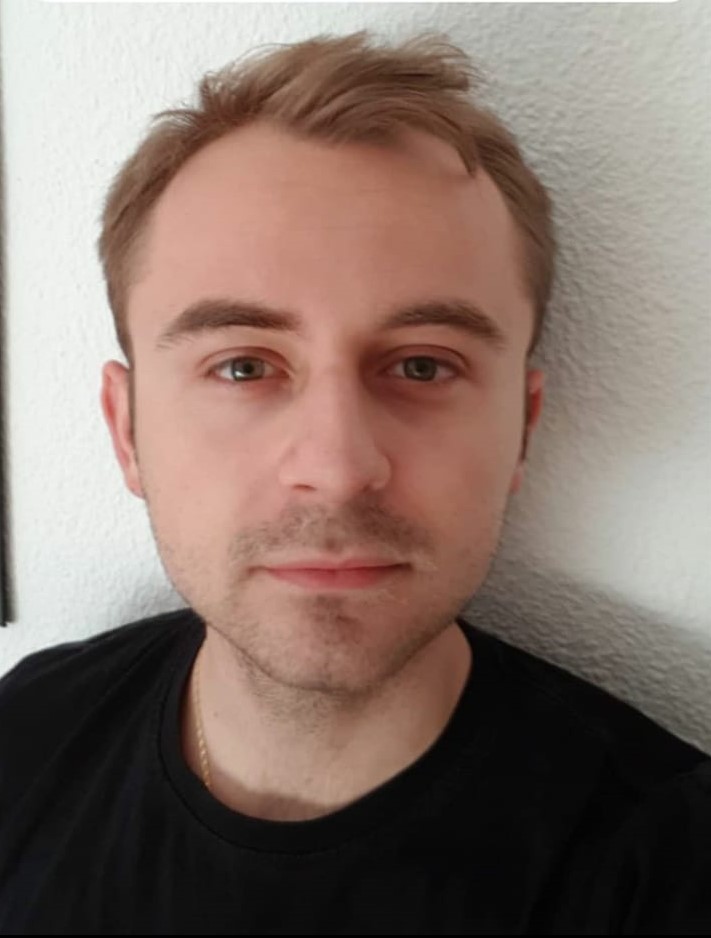}}]{Javier Fumanal Idocin} holds a B.Sc in Computer Science at the University of Zaragoza, Spain and a M.Sc in Data Science and Computer Engineering at the University of Granada, Spain. He is now a PhD Student of the Public University of Navarre, Spain in the department of Statistics, Informatics and Mathematics. His research interests include machine intelligence, fuzzy logic, graph modelling, social networks and BCI Systems.\end{IEEEbiography}
	\begin{IEEEbiography}[{\includegraphics[width=1in,height=1.25in,clip,keepaspectratio]{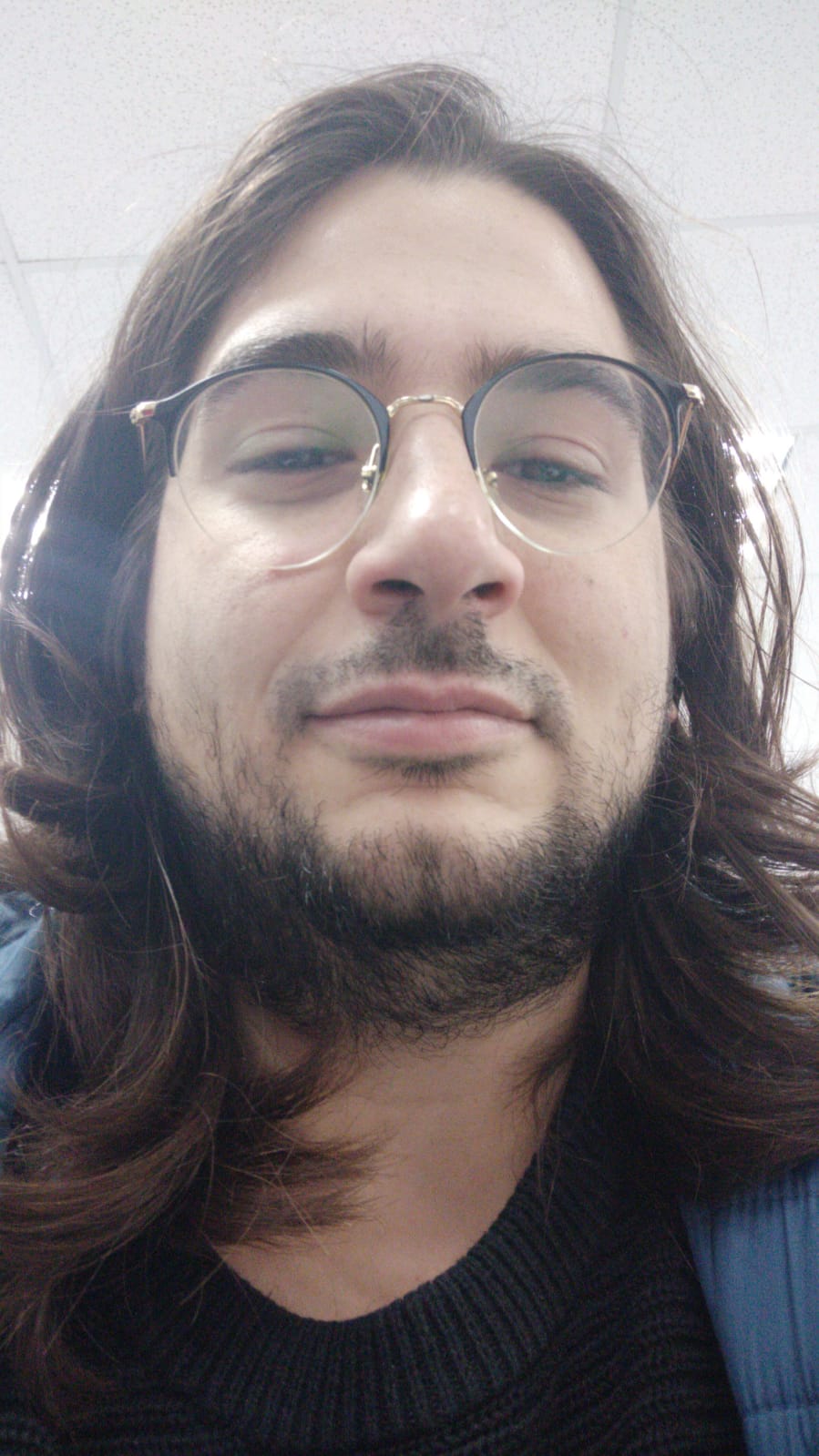}}]{Mattia Delli Priscoli} received a graduation degreee on Computer engineering at the University of Salerno (IT). He is currently Ph.D. student of the DISA-MIS (Department of Business Sciences, Management and Innovation Systems) of the University of
		Salerno (IT). His research interests include artificial and computational intelligence, deep learning and image processing, but he is working on several deep learning tasks, like Microscopy image processing and Distributed neural network training.\end{IEEEbiography}
	\begin{IEEEbiography}[{\includegraphics[width=1in,height=1.25in,clip,keepaspectratio]{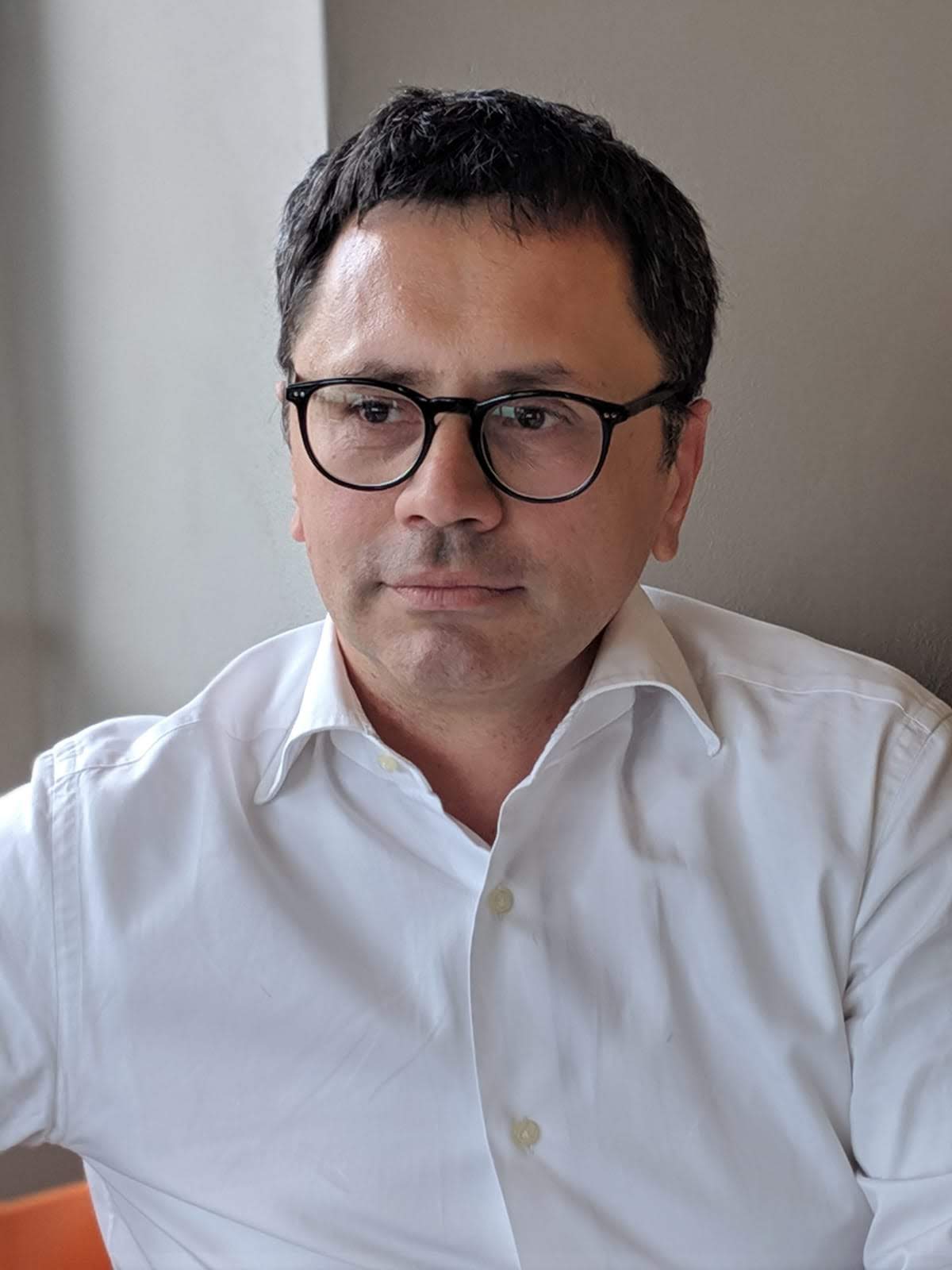}}]{Luigi Troiano}
Luigi Troiano, Ph.D. is Associate Professor of AI, Data Science and Machine Learning at University of Salerno (Italy), Dept. of Management and Innovation Systems. He is coordinator of Computational and Intelligent System Engineering Lab at University of Sannio and NVIDIA Deep Learning Institute University Ambassador. He is chairman of the ISO/JTC 1/SC 42 - “AI and Big Data”, Italian section. His research interests focus on foundational AI and applications to media and finance.
	\end{IEEEbiography}
	\begin{IEEEbiography}[{\includegraphics[width=1in,height=1.25in,clip,keepaspectratio]{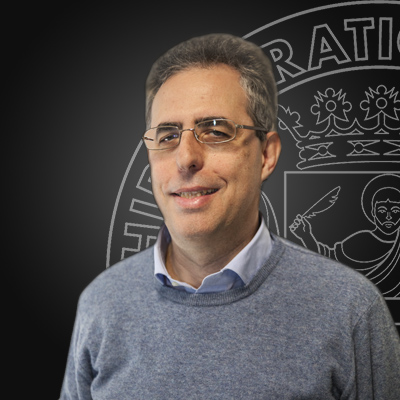}}]{Roberto Tagliaferri}
		Roberto Tagliaferri is full professor in Computer Science at the University of Salerno. He has had courses in Computer Architectures, Artificial and Computational Intelligence, and Bioinformatics for computer scientists and engineers, and biologists.
		He has been co-organizer of international workshops and schools on Neural Nets, Computational Intelligence and Bioinformatics. He has been co-editor of special issues on international journals and of Proceedings of International conferences.
		His research activity has been oriented to Computational Intelligence models and applications in the areas of Astrophysics, Biomedicine, Bioinformatics, and Industrial Applications, with more than 150 publications.
	\end{IEEEbiography}
	
	\begin{IEEEbiography}[{\includegraphics[width=1in,height=1.25in,clip,keepaspectratio]{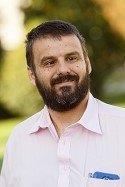}}]{Javier Fernandez}
		received the M.Sc. and Ph.D. degrees in mathematics
		from the University of Zaragoza, Zaragoza, Spain, in 1999 and 2003,
		respectively. He is currently an Associate Lecturer with the Department
		of Statistics, Computer Science and Mathematics, Public University of
		Navarre, Pamplona, Spain.
		He is the author or coauthor of approximately 50 original articles and
		is involved with teaching artificial intelligence and computational
		mathematics for students of the computer sciences. His research
		interests include fuzzy techniques for image processing, fuzzy sets
		theory, interval-valued fuzzy sets theory, aggregation functions, fuzzy
		measures, stability, evolution equation, and unique continuation. He is
		member a member of the Editorial Board of the journal IEEE Transactions
		on Fuzzy Systems.\end{IEEEbiography}
	
	\begin{IEEEbiography}[{\includegraphics[width=1in,height=1.25in,clip,keepaspectratio]{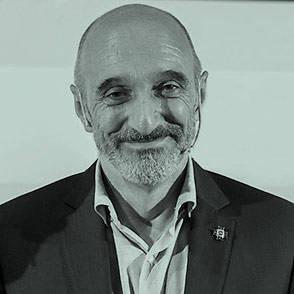}}]{Humberto Bustince}
		(M’08–SM’15) received the Graduate degree in physics
		from the University of Salamanca in 1983 and Ph.D. in mathematics from
		the Public University of Navarra, Pamplona, Spain, in 1994.
		He is a Full Professor of Computer Science and Artificial Intelligence
		in the Public University of Navarra, Pamplona, Spain where he is the
		main researcher of the Artificial Intelligence and Approximate Reasoning
		group, whose main research lines are both theoretical (aggregation
		functions, information and comparison measures, fuzzy sets, and
		extensions) and applied (image processing, classification, machine
		learning, data mining, and big data). He has led 11 I+D public-funded
		research projects, at a national and at a regional level. He is
		currently the main researcher of a project in the Spanish Science
		Program and of a scientific network about fuzzy logic and soft
		computing. He has been in charge of research projects collaborating with
		private companies. He has taken part in two international research
		projects. He has authored more than 210 works, according to Web of
		Science, in conferences and international journals, with around 110 of
		them in journals of the first quartile of JCR. Moreover, five of these
		works are also among the highly cited papers of the last ten years,
		according to Science Essential Indicators of Web of Science.
		Dr. Bustince is the Editor-in-Chief of the online magazine Mathware \&
		Soft Computing of the European Society for Fuzzy Logic and technologies
		and of the Axioms journal. He is an Associated Editor of the IEEE
		Transactions on Fuzzy Systems Journal and a member of the editorial
		board of the Journals Fuzzy Sets and Systems, Information Fusion,
		International Journal of Computational Intelligence Systems and Journal
		of Intelligent \& Fuzzy Systems. He is the coauthor of a monography
		about averaging functions and coeditor of several books. He has
		organized some renowned international conferences such as EUROFUSE 2009
		and AGOP. Honorary Professor at the University of Nottingham, National
		Spanish Computer Science Award in 2019 and EUSFLAT Excellence Research
		Award in 2019.\end{IEEEbiography}
	
	
	
	

\end{document}